\documentclass[10pt,journal,compsoc]{IEEEtran}

\usepackage{afterpage}
\usepackage{stfloats}
\usepackage{cuted}
\usepackage{caption}
\usepackage{graphicx}
\usepackage{adjustbox}
\usepackage{amsmath}
\usepackage{amssymb}
\usepackage{enumitem}
\usepackage{booktabs}
\usepackage{multirow}
\usepackage{url}
\usepackage{subcaption}
\usepackage{array}
\usepackage{tikz}
\usetikzlibrary{shapes,arrows.meta,positioning,fit,calc}
\usepackage[edges]{forest}
\usepackage{ragged2e}
\usepackage[table]{xcolor}
\usepackage{capt-of}

\ifCLASSOPTIONcompsoc
  \usepackage[nocompress]{cite}
\else
  \usepackage{cite}
\fi

\ifCLASSINFOpdf
\else
\fi

\usepackage{epsfig}

\usepackage[pagebackref=true,breaklinks=true,letterpaper=true,colorlinks,bookmarks=false,urlcolor=red,citecolor=cyan]{hyperref}
\usepackage{gensymb,graphics,inputenc}
\usepackage[linesnumbered,algo2e,boxed]{algorithm2e}
\graphicspath{{Figure/}}
\usepackage[figuresright]{rotating}
\usepackage{textcomp,upgreek}
\usepackage{color,mdwlist}
\usepackage{pifont}

\newcommand{\modelname}[1]{MobileVLA-R1 2.0}
\newcommand{\datasetname}[1]{MobileVLA-CoT}
\newcommand{\cmark}{\ding{51}}
\newcommand{\xmark}{\ding{55}}
\usepackage{fontawesome}
\graphicspath{{Figures/}}

\usepackage{caption}
\setitemize{noitemsep,topsep=0pt,parsep=0pt,partopsep=0pt}

\begin{document}
\title{\modelname{}: RL-Enhanced Reasoning for Mobile Robot Control}

\author{Ting Huang$^*$,
        Yue Huang$^*$,
        Zeyu Zhang$^{*\dag}$,
        Shuicheng Yan,~\IEEEmembership{Fellow,~IEEE,}
        Hao Tang$^\ddag$
\IEEEcompsocitemizethanks{
        \IEEEcompsocthanksitem $^*$Equal contribution. $^\dag$Project lead. \protect
        \IEEEcompsocthanksitem $^\ddag$Corresponding author, E-mail: bjdxtanghao@gmail.com. \protect
        \IEEEcompsocthanksitem Ting Huang, Zeyu Zhang and Hao Tang are with the School of Computer Science, Peking University, Beijing 100871, China. \protect
        \IEEEcompsocthanksitem Yue Huang is with South China University of Technology, Guangzhou 510006, China. \protect
        \IEEEcompsocthanksitem Shuicheng Yan is with the School of Computing, National University of Singapore, Singapore 117417. \protect
    }%
}

\markboth{Submitted to IEEE Transactions on Pattern Analysis and Machine Intelligence}%
{Shell \MakeLowercase{\textit{et al.}}: \modelname{}: RL-Enhanced Reasoning for Mobile Robot Control}

\IEEEtitleabstractindextext{%
\justify
\begin{abstract}
Grounding natural-language instructions into reliable and executable actions remains a fundamental challenge for vision-language-action (VLA) systems on mobile robots, due to the persistent gap between high-level semantic reasoning and low-level locomotion and manipulation control.
Existing approaches often rely on implicit reasoning or monolithic action prediction, making it difficult to maintain coherent long-horizon decision making while producing precise and adaptable robot actions.
To address this challenge, we propose \textbf{\modelname{}}, an RL-enhanced VLA framework that explicitly couples structured embodied reasoning with executable mobile robot control.
The framework learns multi-granularity reasoning over embodied trajectories through supervised Chain-of-Thought (CoT) alignment and reinforcement learning, improving reasoning-to-action consistency beyond purely behavioral supervision.
To support both locomotion and manipulation, we further introduce a reasoning-conditioned action decoder that maps multimodal reasoning representations to task-level action targets, which are subsequently translated into embodiment-specific commands by robot controllers.
This design provides a unified perception--reasoning--action interface while decoupling high-level action generation from robot-specific actuation.
We conduct extensive evaluations on language-guided navigation, quadruped control, and humanoid mobile manipulation, covering VLN-CE, QUARD, and real-world deployments on Unitree Go2 and G1 robots.
\modelname{} consistently outperforms strong VLA baselines, achieving an average \textbf{1.6 point} improvement in SR on VLN-CE and a \textbf{10.0 point} improvement in full-task success on real-world G1 mobile manipulation tasks over MobileVLA-R1, while demonstrating robust long-horizon instruction following and closed-loop execution across different robotic platforms.
Code:~\url{https://github.com/AIGeeksGroup/MobileVLA-R1-2.0}.
Website:~\url{https://aigeeksgroup.github.io/MobileVLA-R1-2.0}
\end{abstract}

}

\maketitle

\IEEEdisplaynontitleabstractindextext

\IEEEpeerreviewmaketitle

\section{Introduction}
\IEEEPARstart{V}ision-language-action (VLA) models aim to enable embodied agents to perceive their surroundings, understand natural-language instructions, reason about task objectives, and translate such understanding into executable actions.
For mobile robots, this capability is particularly challenging because semantic decisions must be continuously grounded into physical control under partial observability, sensing uncertainty, actuation noise, and long-horizon task dependencies.
As mobile robots evolve from navigation-oriented platforms toward systems capable of physical interaction, successful execution further requires coordinated locomotion and manipulation.
Achieving reliable grounding from semantic understanding to physical execution therefore remains a fundamental challenge in embodied intelligence.~\cite{anderson2018vision,wang2019reinforced,chaplot2020object,Driess2023PaLMEAE,zitkovich2023rt,huang2022language}

Recent progress in multimodal foundation models has substantially advanced generalist robot policies.
RT-2~\cite{zitkovich2023rt} formulates robot actions as tokens and transfers knowledge from vision-language pretraining to robotic control.
OpenVLA~\cite{kim2024openvla} develops an open-source generalist VLA trained on diverse real-world robot demonstrations, while Octo~\cite{ghosh2024octo} explores large-scale policy pretraining across heterogeneous robotic platforms and action spaces.
More recently, $\pi_0$~\cite{black2024pi0} introduces flow-based continuous action generation, and $\pi_{0.5}$~\cite{black2025pi} further targets open-world generalization and long-horizon behavior.
These advances have considerably improved the generalization and action-generation capabilities of VLA policies, but their decision processes are still predominantly centered on observation-to-action prediction, leaving the intermediate reasoning process largely implicit.

A growing body of work therefore incorporates explicit reasoning into VLA policies.
Inspired by cognitive theories, these efforts aim to introduce System-2-like processing into embodied agents, leveraging the native capacity of System 2 for task decomposition and planning to support task interpretation, intermediate planning, and decision making prior to action execution.
In this work, we use the term embodied reasoning to refer to structured intermediate representations that explicitly encode task interpretation, spatial decisions, and execution strategies.
It serves as a computational approximation of System-2-like deliberation, rather than a full-fledged cognitive System-2 process.
In contrast, most existing VLA policies exhibit System-1-like characteristics: observations and instructions are mapped to actions primarily via implicit representations.
Meanwhile, robot-specific low-level controllers form a System-0-like execution layer that handles fast, embodiment-dependent motor control, removing the burden for the VLA policy to directly learn morphology-specific actuation dynamics.

Embodied Chain-of-Thought (ECoT)~\cite{zawalski2024robotic} introduces structured reasoning over task plans, subtasks, object grounding, and robot states before action prediction.
CoT-VLA~\cite{Zhao2025CoTVLAVC} further explores visual Chain-of-Thought reasoning by predicting intermediate visual goals.
More recent methods investigate tighter reasoning--action coupling: ACoT-VLA~\cite{Zhong2026ACoTVLAAC} introduces action-oriented intermediate reasoning, while dense embodied reasoning approaches~\cite{Li2026TrainingVM} use structured reasoning supervision to shape representations for continuous action generation.
However, existing embodied reasoning approaches primarily focus on producing human-interpretable rationales or intermediate representations, while the explicit connection between System-2-like reasoning and executable System-1/System-0 robot control remains insufficiently explored.
In particular, it remains unclear how deliberative reasoning representations should be converted into compact and executable action abstractions that can be reliably realized by heterogeneous robot controllers.
This challenge highlights the need for an explicit interface that bridges deliberative reasoning with reactive action generation and embodiment-specific execution.

This limitation becomes especially critical when robots must simultaneously reason about semantic goals, spatial constraints, and physical interactions.
The challenge becomes more pronounced in mobile manipulation and humanoid control, where navigation and physical interaction must be coordinated within a single task.
MoManipVLA~\cite{wu2025momanipvla} extends pretrained VLA policies toward mobile manipulation through coordinated base--arm control, while GR00T N1~\cite{Nvidia2025GR00TNA} explores generalist vision-language-action modeling for humanoid robots.
In parallel, reinforcement learning has been increasingly used to improve VLA policies beyond supervised imitation.
MoRE~\cite{zhao2025more} studies reinforcement learning for quadruped VLA control, whereas ReinboT~\cite{zhang2025reinbot} introduces reinforcement learning into vision-language manipulation.
These approaches broaden the capabilities of VLA systems but primarily focus on action generation, embodiment adaptation, or task-level policy optimization.

\begin{figure*}[t]
    \centering
    \small
    \includegraphics[width=\linewidth]{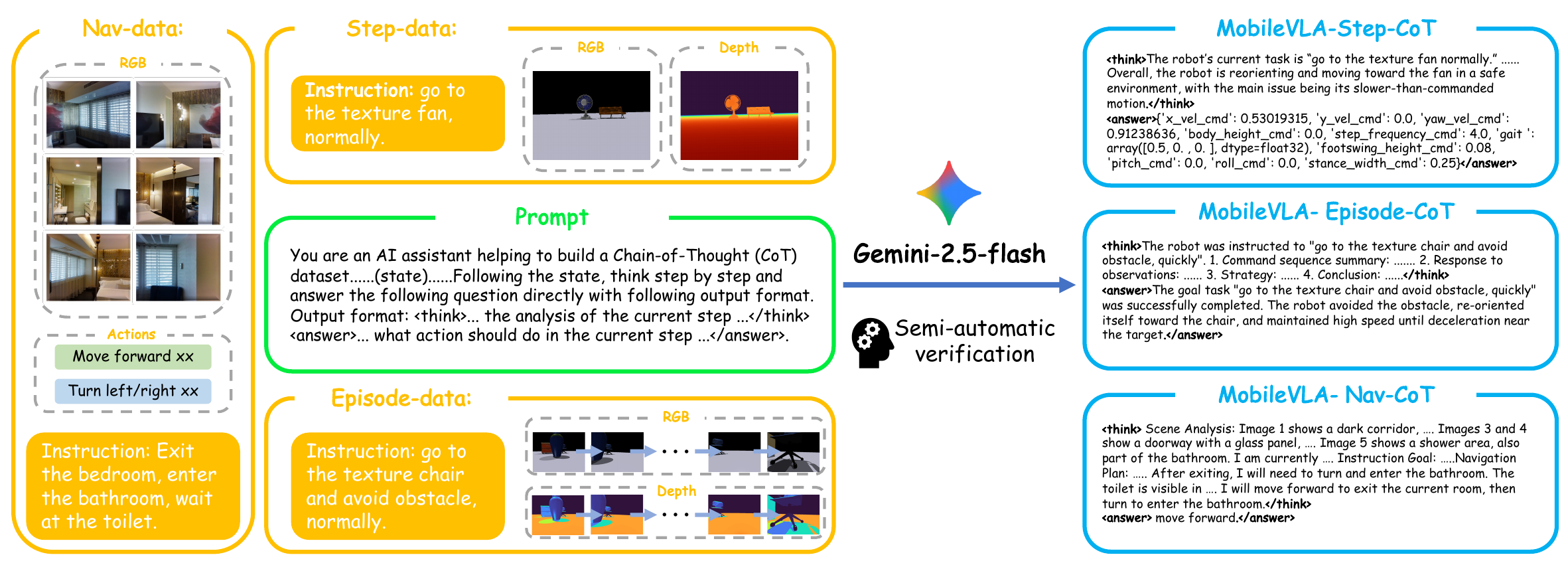}
    \caption{\textbf{Multi-granularity CoT data engine.}
    Given multimodal observations, language instructions, and optional state--action histories, the engine generates episode-, step-, and navigation-level reasoning traces together with executable targets, followed by automatic parsing and semi-automatic quality verification.}
    \label{fig:cot}
    \vspace{-0.6cm}
\end{figure*}

Despite these advances, how to explicitly align structured embodied reasoning with heterogeneous robot controls remains underexplored, particularly for mobile robots that require coordinated locomotion and manipulation.
This motivates a tighter coupling between high-level reasoning and executable action generation.

To this end, we present \textbf{\modelname{}}, an RL-enhanced VLA framework with a reasoning-to-action interface that explicitly grounds embodied reasoning into executable robot behaviors.
Specifically, \modelname{} adopts a System-2-to-System-1-to-System-0 reasoning-execution paradigm: it first performs deliberative embodied reasoning by generating structured Chain-of-Thought (CoT) representations conditioned on multimodal observations and task instructions.
These reasoning representations are subsequently transformed into executable task-level decisions through a reasoning-conditioned action decoder.
Finally, the predicted task-level commands are realized by robot-specific low-level controllers, enabling embodiment-dependent execution while preserving a unified high-level reasoning interface.
To learn this capability, we construct \datasetname{} with complementary episode-, navigation-, and step-level reasoning supervision.
Training proceeds in two stages.
Supervised CoT alignment first establishes structured multimodal reasoning, followed by Group Relative Policy Optimization (GRPO) with execution-aware reward signals to optimize the consistency between reasoning and action generation.
The proposed reasoning-conditioned action decoder serves as an explicit interface between high-level reasoning and physical execution.
Instead of predicting embodiment-specific joint commands, it produces compact task-level locomotion and behavior targets that capture the intended physical behavior.
These targets are subsequently translated into executable commands through robot-specific low-level controllers, separating semantic decision making from embodiment-dependent actuation.
Together, these components establish a unified perception--reasoning--action framework that connects deliberative reasoning with executable control and supports transfer across heterogeneous robot embodiments.

We evaluate \modelname{} across language-guided navigation, quadruped control, and humanoid mobile manipulation.
Our evaluation includes R2R-CE and RxR-CE under the VLN-CE protocol, QUARD for quadruped control, and real-world deployment on a Unitree Go2 robot.
We additionally deploy \modelname{} on a Unitree G1 humanoid robot and evaluate mobile manipulation tasks involving object search, navigation, target approach, grasping, and long-horizon compositions of these skills.
Importantly, no G1-specific trajectories or task annotations are used during training, and the G1 platform is introduced only at evaluation time to assess transfer of the learned reasoning-to-action capability to a humanoid embodiment.
Across these settings, \modelname{} consistently outperforms strong VLA baselines, achieving an average \textbf{1.6 point} improvement in SR on VLN-CE and a \textbf{10.0 point} improvement in full-task success on real-world G1 mobile manipulation tasks over MobileVLA-R1~\cite{huang2025mobilevla}.
These results demonstrate the effectiveness of reinforcement-enhanced reasoning for bridging semantic decision making and executable control across diverse mobile robot tasks.

In summary, our contributions are three-fold:
\begin{itemize}
    \item We propose \textbf{\modelname{}}, an RL-enhanced VLA framework that establishes an explicit reasoning-to-action interface by integrating multi-granularity embodied reasoning, CoT alignment, and reinforcement learning for mobile robot control.

    \item We introduce a reasoning-conditioned action decoder that explicitly maps multimodal observation and reasoning representations into task-level locomotion and behavior predictions, while decoupling semantic action generation from embodiment-specific low-level actuation.

    \item Comprehensive evaluations on VLN-CE, QUARD, and real-world Unitree Go2 and G1 deployments demonstrate effective transfer to humanoid mobile manipulation, with gains of \textbf{1.6 points} in VLN-CE SR and \textbf{10.0 points} in G1 full-task success over MobileVLA-R1.
\end{itemize}

A preliminary version of this work appeared in our ECCV 2026 conference paper~\cite{huang2025mobilevla}.
The present manuscript substantially extends the conference version in both methodology and experimental evaluation in the following six aspects.
(1) We replace the deterministic textual action parsing used in the conference version with a \textbf{reasoning-conditioned action decoder}.
Rather than extracting control commands from generated text through hand-designed parsing rules, the proposed decoder directly maps multimodal observation and intermediate reasoning representations to continuous locomotion commands and discrete task-level behavior primitives, providing an explicit learnable interface between structured reasoning and physical execution.
(2) We introduce an embodiment-decoupled task-level action interface for mobile robot control.
The learned policy predicts semantic task-level actions $(V_x,V_y,\omega,\alpha)$ rather than morphology-specific joint commands, while robot-specific low-level controllers realize these predictions on the physical platform.
This design separates high-level reasoning-to-action prediction from embodiment-specific actuation and enables evaluation on heterogeneous robot platforms without modifying the learned VLA policy.
(3) We substantially extend the real-world evaluation from quadruped navigation and interaction on Unitree Go2 to \textbf{humanoid mobile manipulation on Unitree G1}.
We evaluate long-horizon tasks involving object search, navigation, target approach, grasping, lifting, transporting, and placing under Tabletop, Shelf/Cabinet, and Cluttered settings.
Importantly, the G1 evaluation is conducted without G1-specific trajectories, demonstrations, task annotations, or policy fine-tuning.
(4) We provide substantially more comprehensive analysis of the proposed reasoning-to-action interface.
The journal version includes controlled comparisons between deterministic parsing and learnable decoding, ablations of observation- and reasoning-conditioned decoding, joint analysis of the action decoder and GRPO optimization, and comparisons of alternative decoder architectures.
(5) We broaden the real-world deployment analysis with quantitative efficiency and failure diagnostics.
Beyond success-rate evaluation, the journal version reports end-to-end latency under the hybrid onboard--remote deployment architecture and analyzes episode-level failure modes on Go2, while the G1 evaluation further decomposes failures into grounding, navigation, grasping, manipulation-execution, and low-level control errors.
(6) We provide additional controlled studies of the training and reasoning
components.
These include analyzes of reasoning-supervision granularity, multimodal perception, GRPO reward components and reward-weight sensitivity, rationale sources, policy-optimization objectives, and the interaction between reinforcement optimization and the proposed action decoder.

\section{Related Work}
\noindent\textbf{Language-guided navigation and quadruped VLA.}
Vision-and-language navigation (VLN) studies how embodied agents follow natural-language instructions in visually grounded 3D environments, with R2R~\cite{anderson2018vision} and RxR~\cite{ku2020room} serving as widely used benchmarks.
Advances in pretrained vision-language and 3D multimodal representations~\cite{huang20253d,huang20253dcoca,tang20263d,huang2025dc} have provided increasingly strong semantic and spatial representations for embodied scene understanding.
VLN methods have evolved from sequence prediction~\cite{fried2018speaker,ma2019self} to attention-, memory-, and transformer-based architectures~\cite{hong2020language,zhu2020vision,chen2021history}, and more recently to pretrained vision-language models that improve semantic grounding and generalization to unseen environments~\cite{Hong_2021_CVPR,hao2020towards,qi2021road,rajvanshi2024saynav,Yu2023L3MVNLL,vlnr1,zhang2024uninavid,zhu2025move,liu2025nav}.
In parallel, language-conditioned quadruped policies integrate multimodal perception with locomotion and interaction capabilities, including QUAR-VLA/QUART and their online variants~\cite{ding2024quar,quartonline2025}, as well as generalist quadruped frameworks such as GeRM~\cite{songgerm}.
While these studies have substantially advanced language-guided mobility, they primarily focus on navigation performance or direct action generation.
Our work instead investigates how structured embodied reasoning can be explicitly grounded into executable continuous control.

\noindent\textbf{Generalist and reasoning-enhanced VLA.}
Large-scale multimodal pretraining has enabled generalist VLA models to transfer semantic knowledge from vision-language models to robotic control.
Representative systems such as SayCan~\cite{brohan2023can}, PaLM-E~\cite{Driess2023PaLMEAE}, and RT-2~\cite{zitkovich2023rt} demonstrate the potential of foundation models for language-conditioned robot decision making and action generation.
OpenVLA~\cite{kim2024openvla} and Octo~\cite{ghosh2024octo} further develop generalist policies trained on diverse robot demonstrations and embodiments, while $\pi_0$~\cite{black2024pi0} and $\pi_{0.5}$~\cite{black2025pi} advance continuous action generation and broader task generalization.
Beyond direct observation-to-action prediction, recent work increasingly incorporates explicit intermediate reasoning into VLA policies.
Embodied Chain-of-Thought (ECoT)~\cite{zawalski2024robotic} reasons over plans, subtasks, object grounding, and robot states before action prediction, whereas CoT-VLA~\cite{Zhao2025CoTVLAVC} introduces intermediate visual goals to guide downstream control.
More recent approaches, including ACoT-VLA~\cite{Zhong2026ACoTVLAAC} and dense embodied reasoning methods~\cite{Li2026TrainingVM}, further explore tighter coupling between structured reasoning and continuous action generation.
Despite these advances, reliably translating high-level reasoning into precise and temporally coherent control remains challenging, particularly for tasks involving heterogeneous action spaces.

\noindent\textbf{RL-enhanced VLA and mobile manipulation.}
Reinforcement learning provides a complementary means of improving embodied policies beyond supervised imitation by directly optimizing task- and action-level objectives.
Recent VLA studies have begun to explore this direction: MoRE~\cite{zhao2025more} investigates reinforcement learning for quadruped VLA control, while ReinboT~\cite{zhang2025reinbot} applies reinforcement learning to vision-language manipulation.
Meanwhile, mobile manipulation and humanoid control introduce richer action requirements by coupling mobility with physical interaction.
MoManipVLA~\cite{wu2025momanipvla} adapts pretrained VLA policies to mobile manipulation through coordinated base--arm control, while GR00T N1~\cite{Nvidia2025GR00TNA} develops generalist vision-language-action modeling for humanoid robots with continuous action generation.
These studies substantially broaden the scope of learned robot control, yet the explicit coupling between structured reasoning and coordinated locomotion--manipulation execution remains comparatively underexplored.
In contrast, \modelname{} combines multi-granularity CoT supervision with GRPO-based reasoning-to-action optimization and a reasoning-conditioned action decoder, providing a unified mechanism for grounding structured reasoning into both locomotion and manipulation control.

\begin{table}[t]
\centering
\small
\caption{\textbf{Statistics of source datasets and the synthesized \datasetname{}.}
Nav.'' and Emb-Ctl.'' denote navigation and embodied continuous-control supervision, respectively, while CoT'' indicates synthesized reasoning annotations.
Samples'' denotes the number of instances reported by the corresponding datasets.}
\label{tab:dataset_comparison}
\resizebox{\linewidth}{!}
{\begin{tabular}{lcccc}
\toprule
Dataset                              & Nav. & Emb-Ctl. & CoT & Samples \\
\midrule
R2R~\cite{anderson2018vision}        & \cmark & \xmark & \xmark & 50K \\
RxR~\cite{ku2020room}                & \cmark & \xmark & \xmark & 58K \\
QUARD~\cite{ding2024quar}            & \xmark & \cmark & \xmark & 262K \\
\midrule
\textbf{MobileVLA-CoT-Episode}       & \xmark & \cmark & \cmark & 18K \\
\textbf{MobileVLA-CoT-Step}          & \xmark & \cmark & \cmark & 78K \\
\textbf{MobileVLA-CoT-Nav}           & \cmark & \xmark & \cmark & 38K \\
\bottomrule
\end{tabular}
}
\vspace{-0.5cm}
\end{table}

\section{Datasets}
\label{sec:datasets}
\subsection{Source Datasets}
We construct \datasetname{} from three complementary embodied datasets covering language-guided navigation and continuous robot control.
R2R~\cite{anderson2018vision} provides instruction--trajectory pairs collected in Matterport3D~\cite{matterport3D2017} indoor environments and serves as a standard benchmark for vision-and-language navigation.
RxR~\cite{ku2020room} extends this setting with multilingual and semantically richer instructions, providing stronger supervision for long-horizon instruction grounding.
QUARD~\cite{ding2024quar} complements these navigation datasets with quadruped locomotion and interaction trajectories paired with multimodal observations and executable control targets.
Together, these datasets provide complementary supervision for language-to-trajectory grounding and embodied action generation, forming the basis for constructing multi-granularity reasoning annotations.
Importantly, all reasoning and action supervision used for model training is derived exclusively from R2R, RxR, and QUARD.
No Unitree G1 trajectories, demonstrations, or task-specific annotations are used during dataset construction or model optimization.
The G1 platform is introduced only during real-world evaluation to assess the transferability of the learned reasoning-to-action capability to humanoid mobile manipulation.

\subsection{Multi-Granularity Embodied Reasoning Dataset}
Building on the above source datasets, we construct \datasetname{}, which augments embodied trajectories with structured reasoning traces paired with executable action targets.
Unlike conventional instruction--action supervision that directly associates observations and language instructions with target behaviors, \datasetname{} explicitly introduces intermediate reasoning between task interpretation and physical execution.
This formulation provides structured supervision for learning the reasoning-to-action interface in \modelname{}.

\noindent\textbf{Reasoning granularity.}
\datasetname{} consists of three complementary subsets.
\emph{Episode-level reasoning} summarizes the trajectory outcome, salient observations, and high-level execution strategy over a complete episode.
\emph{Step-level reasoning} explains the action to be executed under the current multimodal observation and state--action history, directly associating local reasoning with executable control.
\emph{Navigation-level reasoning} captures long-horizon spatial decisions that connect a global language instruction to sequential navigation behaviors.
As summarized in Tab.~\ref{tab:dataset_comparison}, the resulting dataset contains 18K episode-level, 78K step-level, and 38K navigation-level samples, totaling 134K reasoning-annotated instances.

\noindent\textbf{Data representation.}
Each sample contains multimodal observations, a natural-language instruction, an optional state--action history, a structured reasoning trace, and an executable target.
The reasoning trace is represented using explicit \texttt{<think>...</think>} delimiters, while the executable output is stored in \texttt{<answer>...</answer>}.
For navigation samples, the target corresponds to a discrete navigation action; for embodied-control samples, it specifies continuous control variables such as translational and angular velocities together with task-specific behaviors.
This unified representation enables the same learning framework to supervise both structured reasoning and executable action generation.

\begin{figure*}[t]
    \centering
    \small
    \includegraphics[width=\linewidth]{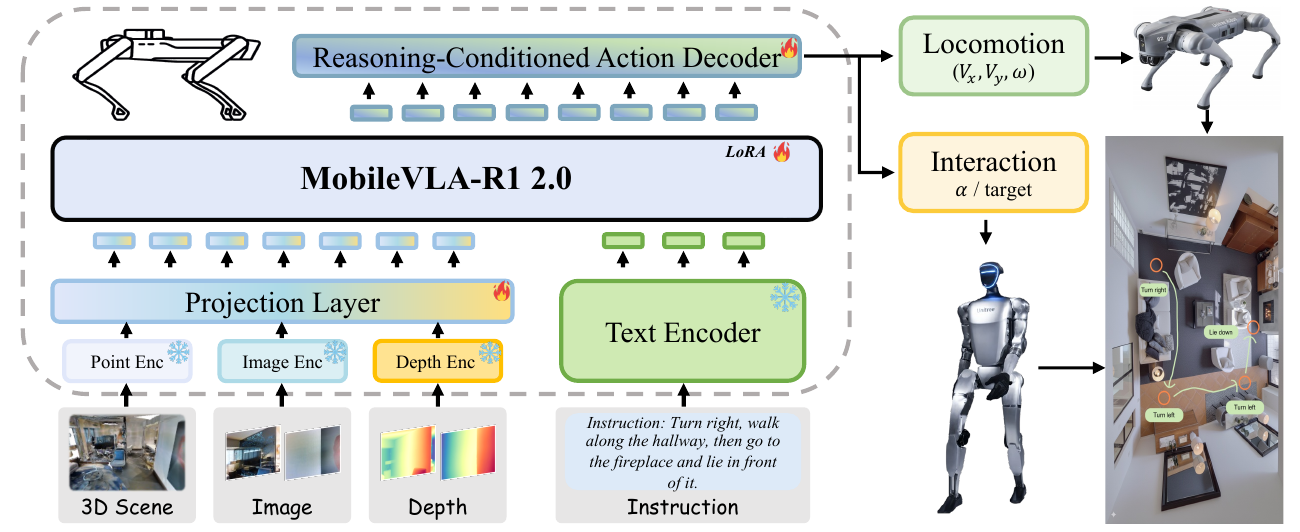}
    \caption{
    \textbf{Overview of \modelname{}.}
    Multimodal observations and language instructions are fused for structured reasoning, which conditions a task-level action decoder to predict locomotion $(V_x,V_y,\omega)$ and a discrete task-level behavior primitive $\alpha$ for mobile robot control.
    }
    \label{fig:model}
    \vspace{-0.6cm}
\end{figure*}

\subsection{CoT Data Engine and Quality Control}
As illustrated in Fig.~\ref{fig:cot}, we develop a model-agnostic CoT data engine for synthesizing structured embodied reasoning at multiple temporal granularities.
The engine takes multimodal observations, natural-language instructions, and optional state--action histories as inputs and applies task-specific prompt templates to elicit a reasoning trace followed by an executable output.
We instantiate the engine with Gemini-2.5-Flash~\cite{comanici2025gemini}; however, the prompting, parsing, and verification procedures are independent of the underlying multimodal model and can be applied to alternative backends.

\noindent\textbf{Quality control.}
We employ a four-stage semi-automatic verification procedure covering format validity, action consistency, safety, and semantic correctness.
Starting from 168K raw generations, we first remove samples with malformed reasoning tags, missing command fields, invalid action ranges, or non-executable outputs.
We subsequently filter unsafe or instruction-irrelevant generations and perform manual inspection to identify hallucinated objects, reasoning--action mismatches, and visual inconsistencies.
The resulting dataset contains 134K validated samples.
Additional details on data-split integrity, filtering criteria, manual verification, and common annotation errors are provided in \textit{App.}~B.

\section{The Proposed Method}
\label{sec:method}

\subsection{Overview and Problem Formulation}
\label{sec:method_overview}
\modelname{} follows a hierarchical \emph{reasoning--execution} paradigm that connects structured embodied reasoning with task-level robot control.
As illustrated in Fig.~\ref{fig:model}, multimodal observations and natural-language instructions are encoded by a VLA backbone to generate structured reasoning, whose internal representations are subsequently mapped to task-level actions by a reasoning-conditioned action decoder.
Training consists of two stages.
We first perform supervised fine-tuning (SFT) on \datasetname{} (Sec.~\ref{sec:datasets}) to align multimodal reasoning with action prediction, and subsequently apply Group Relative Policy Optimization (GRPO)~\cite{shao2024deepseekmath} to further improve reasoning-to-action consistency.

At timestep $t$, the agent receives a multimodal observation
\begin{equation}
s_t=
\left\{
x_t^{\mathrm{rgb}},
x_t^{\mathrm{depth}},
x_t^{\mathrm{pc}}
\right\},
\end{equation}
together with a natural-language instruction $i\in\mathcal{I}$.
The VLA policy generates a structured output
\begin{equation}
o_t
\sim
\pi_{\theta}(\cdot\mid s_t,i),
\end{equation}
following the reasoning--action format
\begin{equation}
o_t=
\texttt{<think>}~r_t~
\texttt{</think><answer>}~u_t~
\texttt{</answer>},
\end{equation}
where $r_t$ denotes the intermediate embodied reasoning trace and $u_t$ provides a structured textual description of the corresponding action intent.

The task-level physical action is represented as
\begin{equation}
\bar{a}_t=
\left[
\mathbf{v}_t,\,
\omega_t,\,
\alpha_t
\right],
\qquad
\mathbf{v}_t=(V_x,V_y),
\end{equation}
where $\mathbf{v}_t$ and $\omega_t$ denote planar translational and yaw velocities, respectively, and $\alpha_t \in \mathcal A_{\rm beh}$ denotes a discrete task-level behavior primitive, such as an interaction, posture, or skill-switching command.
The primitive specifies the semantic behavior to be executed rather than its embodiment-specific joint realization.
These task-level outputs specify the intended physical behavior rather than morphology-specific joint commands and are translated into executable commands by robot-specific low-level controllers.

The structured \texttt{<answer>} output and the action decoder serve complementary roles.
The former provides machine-parsable language supervision and supports format-aware reinforcement optimization, whereas the latter directly maps internal reasoning representations to physical action predictions.

\subsection{Multimodal Reasoning Backbone}
\label{sec:backbone}
As illustrated in Fig.~\ref{fig:model}, \modelname{} adopts a LLaVA-style multimodal architecture~\cite{liu2023visual} initialized from NaVILA~\cite{cheng2025navila}.
Given RGB images, depth maps, point-cloud observations, and a language instruction, modality-specific encoders extract visual and geometric features that are projected into the common hidden space of the VLA backbone.

Let $E_m$ denote the encoder for modality $m\in\{\mathrm{rgb},\mathrm{depth},\mathrm{pc}\}$.
At timestep $t$, the corresponding modality tokens are
\begin{equation}
Z_t^{m}
=
E_m\!\left(x_t^{m}\right)
\in
\mathbb{R}^{N_m\times d_m},
\end{equation}
where $N_m$ and $d_m$ denote the number and dimensionality of modality-specific tokens, respectively.
Each modality is mapped to the common hidden dimension $d$ through a learnable projection layer
$P_m:\mathbb{R}^{d_m}\rightarrow\mathbb{R}^{d}$:
\begin{equation}
\tilde{Z}_t^{m}
=
P_m\!\left(Z_t^{m}\right)
\in
\mathbb{R}^{N_m\times d}.
\end{equation}

The projected multimodal tokens are concatenated with the language tokens:
\begin{equation}
Z_t=
\left[
\tilde{Z}_t^{\mathrm{rgb}};
\tilde{Z}_t^{\mathrm{depth}};
\tilde{Z}_t^{\mathrm{pc}};
Z^{\mathrm{text}}
\right]
\in
\mathbb{R}^{N_{\mathrm{in}}\times d},
\end{equation}
where
$N_{\mathrm{in}}
=
N_{\mathrm{rgb}}
+
N_{\mathrm{depth}}
+
N_{\mathrm{pc}}
+
N_{\mathrm{text}}$
is the total number of input tokens.
Modality-type embeddings are added before backbone processing to preserve the source identity of the multimodal tokens.

The resulting multimodal-language sequence is contextualized by the VLA backbone:
\begin{equation}
H_t^{\mathrm{ctx}}
=
F_{\theta}\!\left(Z_t\right)
\in
\mathbb{R}^{N_{\mathrm{in}}\times d},
\end{equation}
where $F_{\theta}$ denotes the NaVILA-initialized backbone.
We denote by
$H_t^{\mathrm{obs}}
=
H_t^{\mathrm{ctx}}[\mathcal{I}_{\mathrm{obs}}]$
the contextual states associated with the RGB, depth, and point-cloud tokens, where $\mathcal{I}_{\mathrm{obs}}$ indexes the corresponding observation tokens.

Conditioned on the multimodal-language context, the backbone autoregressively generates the structured reasoning sequence.
The hidden states associated with the generated reasoning tokens are denoted by $H_t^{\mathrm{rea}}\in\mathbb{R}^{L_r\times d}$, where $L_r$ is the reasoning-sequence length.
Together, $H_t^{\mathrm{obs}}$ provides perceptual and geometric context, whereas $H_t^{\mathrm{rea}}$ provides reasoning context for the action decoder described in Sec.~\ref{sec:action_decoder}.

During supervised alignment, the modality encoders remain frozen, whereas the multimodal projection layers, LoRA parameters of the VLA backbone, and the reasoning-conditioned action decoder are optimized.
Additional architectural configurations and modality-specific token budgets are provided in Sec.~\ref{sec:supp_arch_details}.

\begin{figure*}[t]
    \centering
    \small
    \includegraphics[width=\linewidth]{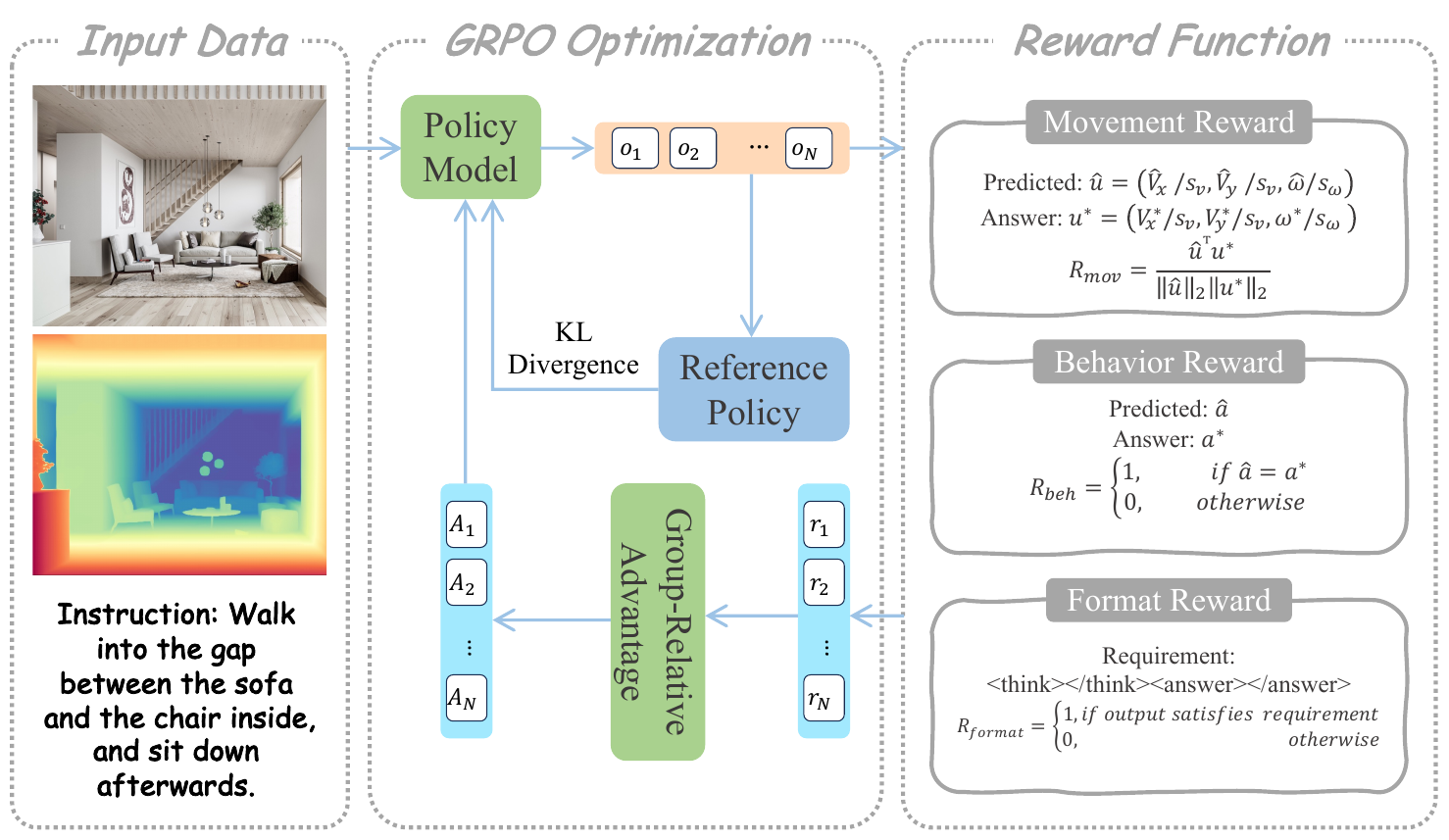}
    \caption{\textbf{GRPO-based reasoning-to-action optimization.}
    For each multimodal input, the policy samples multiple structured outputs whose induced actions are evaluated by movement, behavior, and format rewards.
    Group-relative advantages, together with KL regularization to a frozen reference policy, are then used for offline policy optimization.
    }
    \label{fig:rl}
    \vspace{-0.6cm}
\end{figure*}

\subsection{Reasoning-Conditioned Action Decoder}
\label{sec:action_decoder}
Structured CoT exposes the intermediate decision process of the VLA policy, but directly parsing generated text into control commands provides only an implicit connection between reasoning and physical execution.
We therefore introduce a learnable \emph{reasoning-conditioned action decoder} that explicitly maps reasoning and observation representations to task-level robot actions.

\noindent\textbf{Reasoning-conditioned action representation.}
Given $H_t^{\mathrm{obs}}$ and $H_t^{\mathrm{rea}}$, we construct the joint contextual sequence
\begin{equation}
K_t=
\left[
H_t^{\mathrm{obs}};
H_t^{\mathrm{rea}}
\right]
\in
\mathbb{R}^{(N_{\mathrm{obs}}+L_r)\times d}.
\end{equation}
To capture the different control semantics of locomotion and behavior, we introduce two learnable action queries $q_{\mathrm{loc}},q_{\mathrm{beh}}\in\mathbb{R}^{1\times d}$.
For branch $b\in\{\mathrm{loc},\mathrm{beh}\}$, the corresponding reasoning-conditioned action representation is obtained through cross-attention:
\begin{equation}
h_t^{b}
=
\operatorname{CrossAttn}
\left(
q_b,
K_t,
K_t
\right)
\in
\mathbb{R}^{d}.
\end{equation}
The joint context provides the keys and values, while the corresponding action query selectively aggregates information relevant to each control branch.

\noindent\textbf{Locomotion prediction.}
The locomotion representation is mapped to continuous planar motion through a lightweight regression head:
\begin{equation}
\left[
\hat{\mathbf{v}}_t,
\hat{\omega}_t
\right]
=
D_{\mathrm{loc}}
\left(
h_t^{\mathrm{loc}}
\right),
\end{equation}
where $\hat{\mathbf{v}}_t=(\hat{V}_{x,t},\hat{V}_{y,t})$ and $\hat{\omega}_t$ denote the predicted translational and yaw velocities, respectively.

\noindent\textbf{Behavior prediction.}
The behavior representation predicts a distribution over the predefined task-level behavior set $\mathcal{A}_{\mathrm{beh}}$:
\begin{equation}
p_{\phi}
\left(
\alpha_t
\mid
h_t^{\mathrm{beh}}
\right)
=
\operatorname{Softmax}
\left(
D_{\mathrm{beh}}
\left(
h_t^{\mathrm{beh}}
\right)
\right).
\end{equation}
At inference time, the highest-probability behavior primitive is selected, yielding
$\hat{\bar{a}}_t=
[\hat{\mathbf{v}}_t,\hat{\omega}_t,\hat{\alpha}_t]$.
We collectively denote the parameters of the cross-attention module, action queries, and prediction heads by $\phi$.

Unlike deterministic command parsing, the proposed decoder establishes a learnable mapping from structured reasoning representations to task-level physical actions.
The textual \texttt{<answer>} output is retained for structured language supervision and format-aware optimization, whereas physical execution is driven by the decoder predictions.

\noindent\textbf{Embodiment-specific execution.}
The decoder predicts task-level actions rather than morphology-specific joint commands. For locomotion, $(V_x,V_y,\omega)$ specifies the desired planar motion and is executed by the robot-specific locomotion controller.
The discrete behavior output $\alpha_t$ specifies a task-level behavior primitive. On the G1 platform, manipulation-related primitives such as reaching, grasping, lifting, transporting, and placing are executed by fixed controller-side manipulation routines rather than being predicted as joint-level trajectories by the VLA policy.
Consequently, the learned policy determines \emph{what} task-level behavior to execute, while the embodiment-specific controller determines \emph{how} that behavior is realized on the physical robot.
No G1-specific trajectories or demonstrations are used to optimize \modelname{}; the G1 platform and its fixed low-level controllers are introduced only during evaluation.

\subsection{Supervised Reasoning and Action Alignment}
Before reinforcement optimization, we perform supervised fine-tuning to establish structured embodied reasoning and align the resulting reasoning representations with task-level action prediction.

Across the supervised stages, the structured output is optimized autoregressively.
Given a target sequence $o_t^{*}=\{y_1^{*},\ldots,y_L^{*}\}$, we define
\begin{equation}
\mathcal{L}_{\mathrm{CoT}}
=
-
\sum_{\ell=1}^{L}
\log
p_{\theta}
\left(
y_{\ell}^{*}
\mid
Z_t,
y_{<\ell}^{*}
\right).
\end{equation}
This objective supervises both the \texttt{<think>} reasoning trace and the structured \texttt{<answer>} output.

\noindent\textbf{Long-horizon reasoning alignment.}
We first fine-tune \modelname{} on MobileVLA-CoT-Episode and MobileVLA-CoT-Nav.
Episode-level supervision captures trajectory-level task progression and high-level execution strategies, whereas navigation-level supervision associates language instructions with sequential spatial decisions.
Together, these data establish multimodal instruction grounding and temporally coherent reasoning over long-horizon embodied trajectories.

\noindent\textbf{Reasoning-to-action alignment.}
We subsequently fine-tune on MobileVLA-CoT-Step, which provides step-level associations between local reasoning and executable task-level actions.
At this stage, the reasoning-conditioned action decoder is jointly optimized with $\mathcal{L}_{\mathrm{CoT}}$.

For continuous locomotion, we use
\begin{equation}
\mathcal{L}_{\mathrm{loc}}
=
\left\|
\hat{\mathbf{v}}_t-\mathbf{v}_t^{*}
\right\|_{1}
+
\lambda_{\omega}
\left|
\hat{\omega}_t-\omega_t^{*}
\right|,
\end{equation}
where $\mathbf{v}_t^{*}$ and $\omega_t^{*}$ denote the target translational and yaw velocities, respectively.
For the discrete behavior primitive, we use
\begin{equation}
\mathcal{L}_{\mathrm{beh}}
=
-
\log
p_{\phi}
\left(
\alpha_t^{*}
\mid
h_t^{\mathrm{beh}}
\right),
\end{equation}
where $\alpha_t^{*}\in\mathcal{A}_{\mathrm{beh}}$ denotes the target behavior primitive.

Since not all training samples provide both forms of action supervision, we use binary indicators $m_{\mathrm{loc}},m_{\mathrm{beh}}\in\{0,1\}$ to activate the corresponding losses only when valid targets are available.
The overall supervised objective is
\begin{equation}
\mathcal{L}_{\mathrm{SFT}}
=
\mathcal{L}_{\mathrm{CoT}}
+
\lambda_{\mathrm{loc}}
m_{\mathrm{loc}}
\mathcal{L}_{\mathrm{loc}}
+
\lambda_{\mathrm{beh}}
m_{\mathrm{beh}}
\mathcal{L}_{\mathrm{beh}},
\end{equation}
where $\lambda_{\mathrm{loc}}$ and $\lambda_{\mathrm{beh}}$ balance locomotion and behavior supervision.
The joint objective establishes an explicit correspondence between intermediate reasoning representations and task-level physical actions.

\begin{table*}[tb]
\centering
\small
\caption{\textbf{Comparison on VLN-CE Val-Unseen~\cite{krantz2020beyond,ku2020room}.}
We report standard metrics on the R2R-CE and RxR-CE val-unseen splits.
``S.RGB'', ``Pano.'', and ``Odo.'' denote single-view RGB, panoramic observations, and odometry, respectively.
$^*$ indicates the use of a simulator-pretrained waypoint predictor~\cite{krantz2020beyond}.
}
\label{tab:vlnce}
\resizebox{0.9\linewidth}{!}{
\begin{tabular}{lcccccccccccc}
    \toprule
    \multirow{2}[2]{*}{Method} & \multicolumn{4}{c}{Observation} & \multicolumn{4}{c}{R2R-CE Val-Unseen} & \multicolumn{4}{c}{RxR-CE Val-Unseen} \\
    \cmidrule(lr){2-5} \cmidrule(lr){6-9} \cmidrule(lr){10-13}
    & S.RGB & Pano. & Depth & Odom. & NE$\downarrow$ & OS$\uparrow$ & SR$\uparrow$ & SPL$\uparrow$ & NE$\downarrow$ & SR$\uparrow$ & SPL$\uparrow$ & nDTW$\uparrow$ \\
    \midrule
    CMA$^{*}$~\cite{hong2022bridging}           & \xmark & \cmark & \cmark & \cmark & 6.20 & 52.0 & 41.0 & 36.0 & 8.76 & 26.5 & 22.1 & 47.0 \\
    Sim2Sim$^{*}$~\cite{krantz2022sim}          & \xmark & \cmark & \cmark & \cmark & 6.07 & 52.0 & 43.0 & 36.0 & - & - & - & - \\
    GridMM$^{*}$~\cite{wang2023gridmm}          & \xmark & \cmark & \cmark & \cmark & 5.11 & 61.0 & 49.0 & 41.0 & - & - & - & - \\
    Ego$^{2}$-Map$^{*}$~\cite{hong2023learning} & \xmark & \cmark & \cmark & \cmark & 5.54 & 56.0 & 47.0 & 41.0 & - & - & - & - \\
    DreamWalker$^{*}$~\cite{wang2023dreamwalker}& \xmark & \cmark & \cmark & \cmark & 5.53 & 59.0 & 49.0 & 44.0 & - & - & - & - \\
    Reborn$^{*}$~\cite{an20221st}               & \xmark & \cmark & \cmark & \cmark & 5.40 & 57.0 & 50.0 & 46.0 & 5.98 & 48.6 & 42.0 & 63.3 \\
    ETPNav$^{*}$~\cite{an2024etpnav}            & \xmark & \cmark & \cmark & \cmark & 4.71 & 65.0 & 57.0 & 49.0 & 5.64 & 54.7 & 44.8 & 61.9 \\
    HNR$^{*}$~\cite{wang2024lookahead}          & \xmark & \cmark & \cmark & \cmark & 4.42 & 67.0 & 61.0 & 51.0 & 5.50 & 56.3 & 46.7 & 63.5 \\
    \midrule
    AG-CMTP~\cite{chen2021topological}          & \xmark & \cmark & \cmark & \cmark & 7.90 & 39.0 & 23.0 & 19.0 & - & - & - & - \\
    R2R-CMTP~\cite{chen2021topological}         & \xmark & \cmark & \cmark & \cmark & 7.90 & 38.0 & 26.0 & 22.0 & - & - & - & - \\
    InstructNav~\cite{long2024instructnav}      & \xmark & \cmark & \cmark & \cmark & 6.89 & -    & 31.0 & 24.0 & - & - & - & - \\
    LAW~\cite{raychaudhuri2021language}         & \cmark & \xmark & \cmark & \cmark & 6.83 & 44.0 & 35.0 & 31.0 & 10.90 & 8.0 & 8.0 & 38.0 \\
    CM2~\cite{georgakis2022cross}               & \cmark & \xmark & \cmark & \cmark & 7.02 & 41.0 & 34.0 & 27.0 & - & - & - & - \\
    WS-MGMap~\cite{chen2022weakly}              & \cmark & \xmark & \cmark & \cmark & 6.28 & 47.0 & 38.0 & 34.0 & - & - & - & - \\
    AO-Planner~\cite{chen2025affordances}       & \xmark & \cmark & \cmark & \xmark & 5.55 & 59.0 & 47.0 & 33.0 & 7.06 & 43.3 & 30.5 & 50.1 \\
    Seq2Seq~\cite{krantz2020beyond}             & \cmark & \xmark & \cmark & \xmark & 7.77 & 37.0 & 25.0 & 22.0 & 12.10 & 13.9 & 11.9 & 30.8 \\
    CMA~\cite{krantz2020beyond}                 & \cmark & \xmark & \cmark & \xmark & 7.37 & 40.0 & 32.0 & 30.0 & - & - & - & - \\
    NaVid~\cite{zhang2024navid}                 & \cmark & \xmark & \xmark & \xmark & 5.47 & 49.0 & 37.0 & 35.0 & - & - & - & - \\
    Uni-NaVid~\cite{zhang2024uninavid}          & \cmark & \xmark & \xmark & \xmark & 5.58 & 53.5 & 47.0 & 42.7 & 6.24 & 48.7 & 40.9 & - \\
    NaVILA~\cite{cheng2025navila}               & \cmark & \xmark & \xmark & \xmark & 5.22 & 62.5 & 54.0 & 49.0 & 6.77 & 49.3 & 44.0 & 58.8 \\
    VLN-R1~\cite{vlnr1}                         & \cmark & \xmark & \xmark & \xmark & 7.00 & 41.2 & 30.2 & 21.8 & 9.10 & 22.7 & 17.6 & -    \\
    OctoNav~\cite{gao2025octonav}               & \cmark & \xmark & \xmark & \xmark & -    & 42.9 & 37.1 & 33.6 & -    & -    & -    & -    \\
    StreamVLN~\cite{wei2025streamvln}           & \cmark & \xmark & \xmark & \xmark & 4.98 & 64.2 & 56.9 & 51.9 & 6.22 & 52.9 & 46.0 & 61.9 \\
    CorrectNav~\cite{yu2025correctnav}          & \cmark & \xmark & \xmark & \xmark & 4.24 & 67.5 & 65.1 & 62.3 & 4.09 & 69.3 & 63.3 & 75.2 \\
    \midrule
    MobileVLA-R1~\cite{huang2025mobilevla}                                & \cmark & \xmark & \cmark & \xmark & 4.05 & 69.7 & 68.3 & 65.2 & 3.92 & 71.5 & 66.8 & 76.1 \\
    \textbf{\modelname{} (Ours)}                & \cmark & \xmark & \cmark & \xmark & \textbf{3.86} & \textbf{71.2} & \textbf{69.8} & \textbf{66.9} & \textbf{3.71} & \textbf{73.1} & \textbf{68.5} & \textbf{77.6} \\
    \bottomrule
\end{tabular}
}
\vspace{-0.6cm}
\end{table*}

\subsection{GRPO-Based Reasoning-to-Action Optimization}
Following supervised alignment, we employ offline GRPO~\cite{shao2024deepseekmath} to further optimize reasoning-to-action consistency.
GRPO constructs relative advantages from multiple candidate outputs generated for the same input without requiring an additional learned value model.
The optimization is performed on a fixed embodied dataset without environment interaction or online robot adaptation.

During GRPO, the reasoning-conditioned action decoder $D_{\phi}$ is kept fixed, while the trainable VLA policy is optimized using rewards derived from the task-level actions induced by sampled reasoning trajectories.
The decoder therefore provides a fixed reasoning-to-action interface for reward evaluation and receives no gradient from the GRPO objective.

\noindent\textbf{Group sampling and action decoding.}
Given a multimodal observation--instruction pair $(s_t,i)$, the sampling policy $\pi_{\theta_{\mathrm{old}}}$ generates $N$ candidate outputs $\{o_{t,j}\}_{j=1}^{N}$.
For each candidate, we extract its reasoning representation $H_{t,j}^{\mathrm{rea}}$ and obtain the corresponding task-level action through the fixed decoder:
\begin{equation}
\hat{\bar{a}}_{t,j}
=
D_{\phi}
\left(
H_t^{\mathrm{obs}},
H_{t,j}^{\mathrm{rea}}
\right)
=
\left[
\hat{\mathbf{v}}_{t,j},
\hat{\omega}_{t,j},
\hat{\alpha}_{t,j}
\right].
\end{equation}
The sampled output and its induced action are evaluated using complementary movement, behavior, and format rewards.

\noindent\textbf{Movement reward.}
We evaluate locomotion consistency in a normalized command space:
\begin{equation}
\begin{aligned}
\hat{\mathbf{u}}_{t,j}
&=
\left(
\frac{\hat{V}_{x,t,j}}{s_v},
\frac{\hat{V}_{y,t,j}}{s_v},
\frac{\hat{\omega}_{t,j}}{s_{\omega}}
\right),\\
\mathbf{u}_{t}^{*}
&=
\left(
\frac{V_{x,t}^{*}}{s_v},
\frac{V_{y,t}^{*}}{s_v},
\frac{\omega_{t}^{*}}{s_{\omega}}
\right),
\end{aligned}
\end{equation}
where $s_v$ and $s_{\omega}$ are fixed normalization factors for translational and angular velocities.
The movement reward is defined by cosine similarity:
\begin{equation}
R_{\mathrm{mov}}(o_{t,j})
=
\frac{
\hat{\mathbf{u}}_{t,j}^{\top}\mathbf{u}_{t}^{*}
}{
\|\hat{\mathbf{u}}_{t,j}\|_2
\|\mathbf{u}_{t}^{*}\|_2
}.
\end{equation}

\noindent\textbf{Behavior reward.}
For the discrete task-level behavior primitive, we use an exact-match reward:
\begin{equation}
R_{\mathrm{beh}}(o_{t,j})
=
\mathbb{I}
\left[
\hat{\alpha}_{t,j}
=
\alpha_t^{*}
\right].
\end{equation}

\noindent\textbf{Format reward.}
To preserve the structured reasoning interface, we define
\begin{equation}
R_{\mathrm{fmt}}(o_{t,j})
=
\mathbb{I}\!\left[
o_{t,j}\in\mathcal{F}
\right],
\end{equation}
where $\mathcal{F}$ denotes outputs conforming to the \texttt{<think>...</think><answer>...</answer>} format.

\noindent\textbf{Composite reward and group-relative advantage.}
The overall reward is
\begin{equation}
r_{t,j}
=
\lambda_{\mathrm{mov}}
R_{\mathrm{mov}}(o_{t,j})
+
\lambda_{\mathrm{beh}}
R_{\mathrm{beh}}(o_{t,j})
+
\lambda_{\mathrm{fmt}}
R_{\mathrm{fmt}}(o_{t,j}),
\end{equation}
where
$\lambda_{\mathrm{mov}}$,
$\lambda_{\mathrm{beh}}$, and
$\lambda_{\mathrm{fmt}}$
control the contributions of the three terms.

The group-relative advantage is computed as
\begin{equation}
\hat{A}_{t,j}
=
\frac{
r_{t,j}-\bar{r}_t
}{
\sigma_{r,t}+\epsilon
},
\end{equation}
where $\bar{r}_t$ and $\sigma_{r,t}$ denote the mean and standard deviation of the $N$ rewards within the corresponding group, respectively.

\begin{figure*}[!t]
    \centering
    \includegraphics[width=0.8\linewidth]{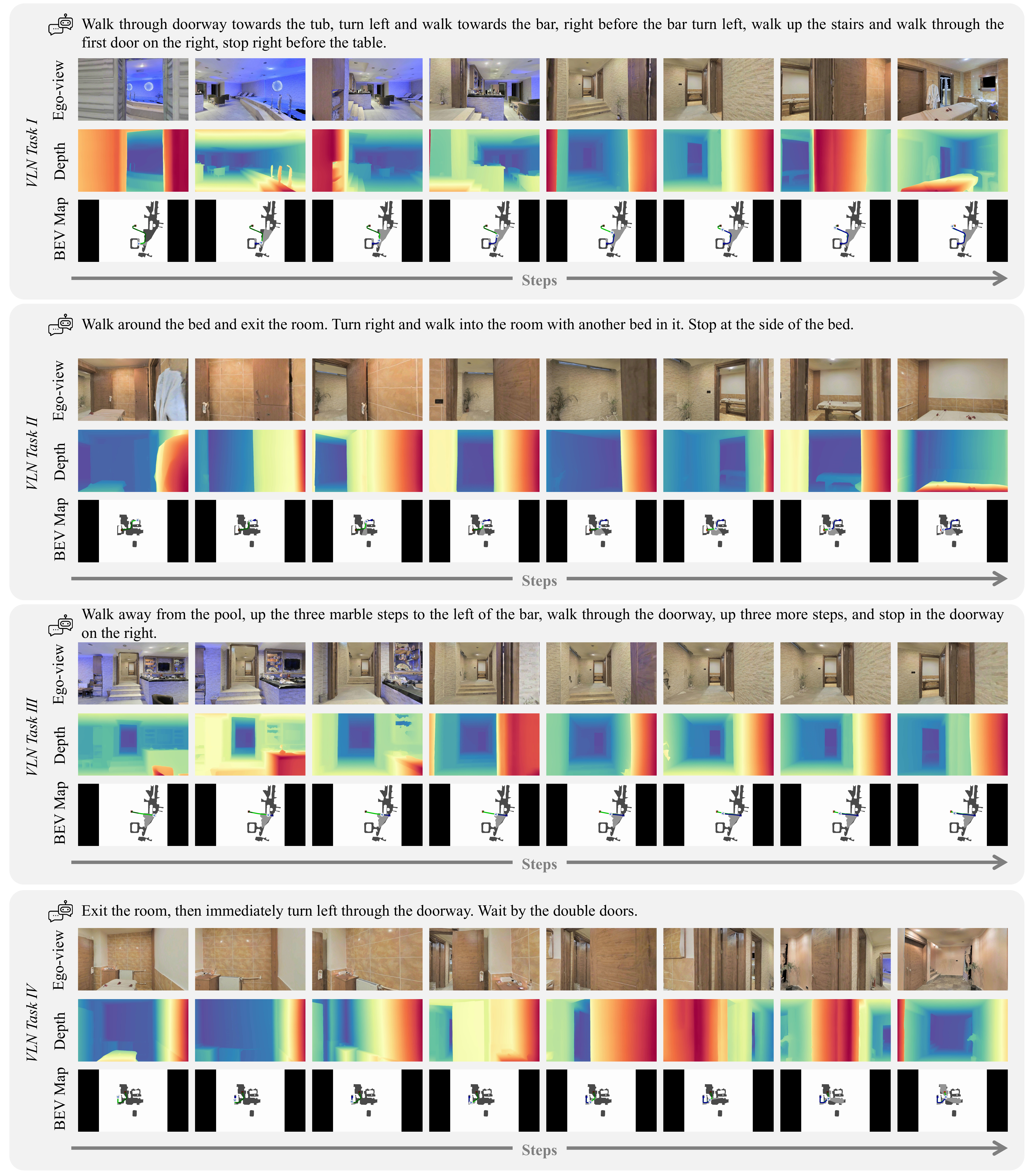}
    \caption{\textbf{Qualitative results on R2R-CE.} \modelname{} executes navigation instructions in the simulator.
    }
    \label{fig:simulator_vln_1}
    \vspace{-0.6cm}
\end{figure*}
\noindent\textbf{GRPO objective.}
Let $o_{t,j}=\{y_{t,j,1},\ldots,y_{t,j,L_j}\}$ denote the $j$-th sampled sequence.
For token $\ell$, the policy ratio is
\begin{equation}
\rho_{t,j,\ell}(\theta)
=
\frac{
\pi_{\theta}
\left(
y_{t,j,\ell}
\mid
c_{t,j,\ell}
\right)
}{
\pi_{\theta_{\mathrm{old}}}
\left(
y_{t,j,\ell}
\mid
c_{t,j,\ell}
\right)
},
\end{equation}
where $c_{t,j,\ell}=(s_t,i,y_{t,j,<\ell})$ denotes the autoregressive generation context.
The clipped GRPO objective is
\begin{equation}
\begin{aligned}
J_{\mathrm{GRPO}}(\theta)
=
\mathbb{E}_{(s_t,i)\sim\mathcal{D}}
\Bigg[
\frac{1}{N}
\sum_{j=1}^{N}
\frac{1}{L_j}
\sum_{\ell=1}^{L_j}
\Big(
\min\{
\rho_{t,j,\ell}(\theta)\hat{A}_{t,j},\\
\operatorname{clip}
(
\rho_{t,j,\ell}(\theta),
1-\epsilon_c,
1+\epsilon_c
)
\hat{A}_{t,j}
\}\\
-
\beta
D_{\mathrm{KL}}
\big(
\pi_{\theta}(\cdot\mid c_{t,j,\ell})
\|
\pi_{\mathrm{ref}}(\cdot\mid c_{t,j,\ell})
\big)
\Big)
\Bigg].
\end{aligned}
\end{equation}
Here, $\pi_{\mathrm{ref}}$ denotes the frozen reference policy, $\epsilon_c$ is the clipping threshold, and $\beta$ controls the strength of KL regularization.
The objective favors reasoning trajectories that induce more accurate task-level actions while constraining excessive deviation from the reference policy.

\begin{figure*}[!t]
    \centering
    \includegraphics[width=0.8\linewidth]{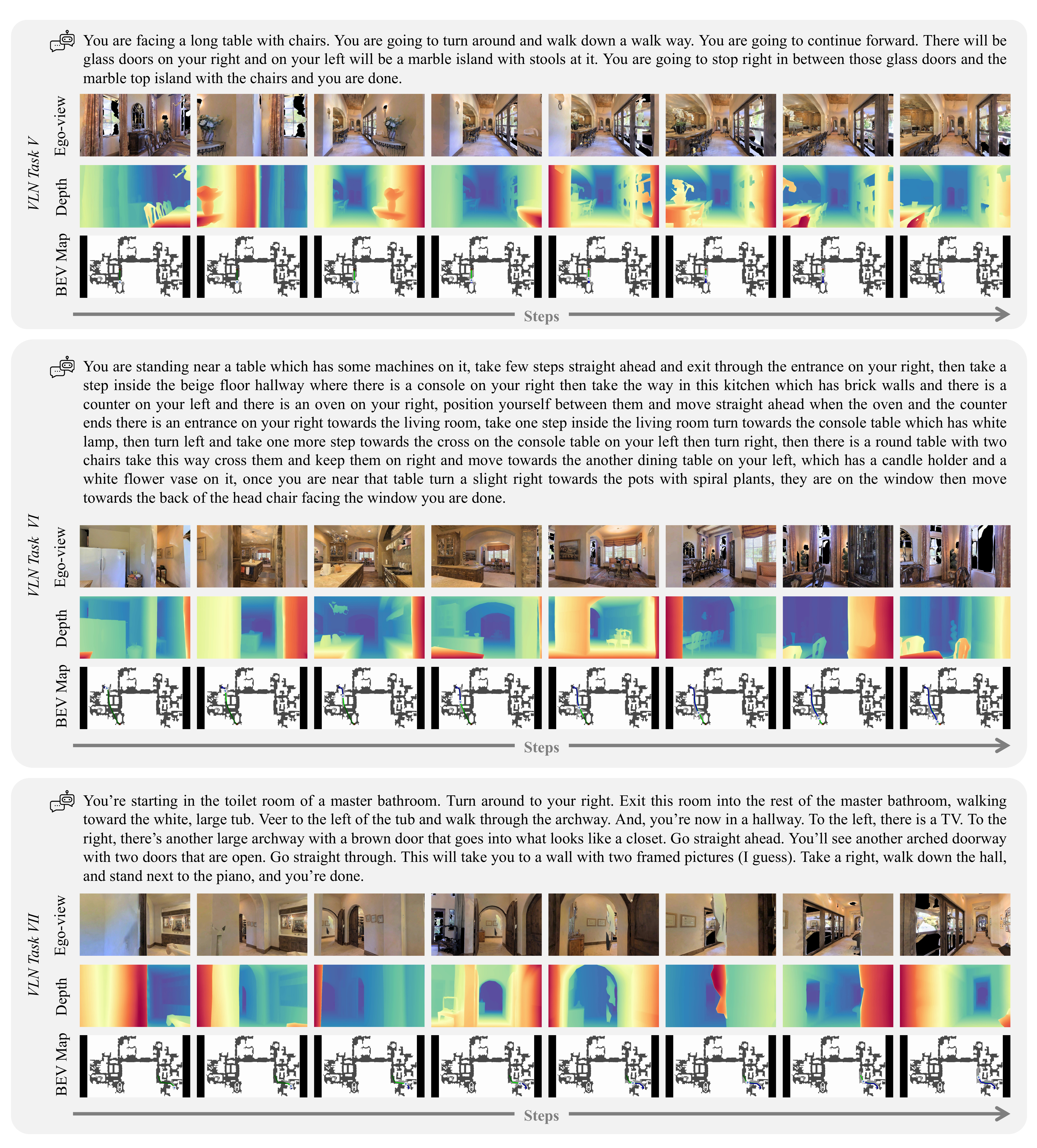}
    \caption{\textbf{Qualitative results on RxR-CE.} \modelname{} follows multilingual instructions with complex spatial semantics, maintaining coherent reasoning and stable motion across unseen environments.
    }
    \label{fig:simulator_vln_2}
    \vspace{-0.6cm}
\end{figure*}

\begin{figure}[t]
    \centering
    \small
    \includegraphics[width=\linewidth]{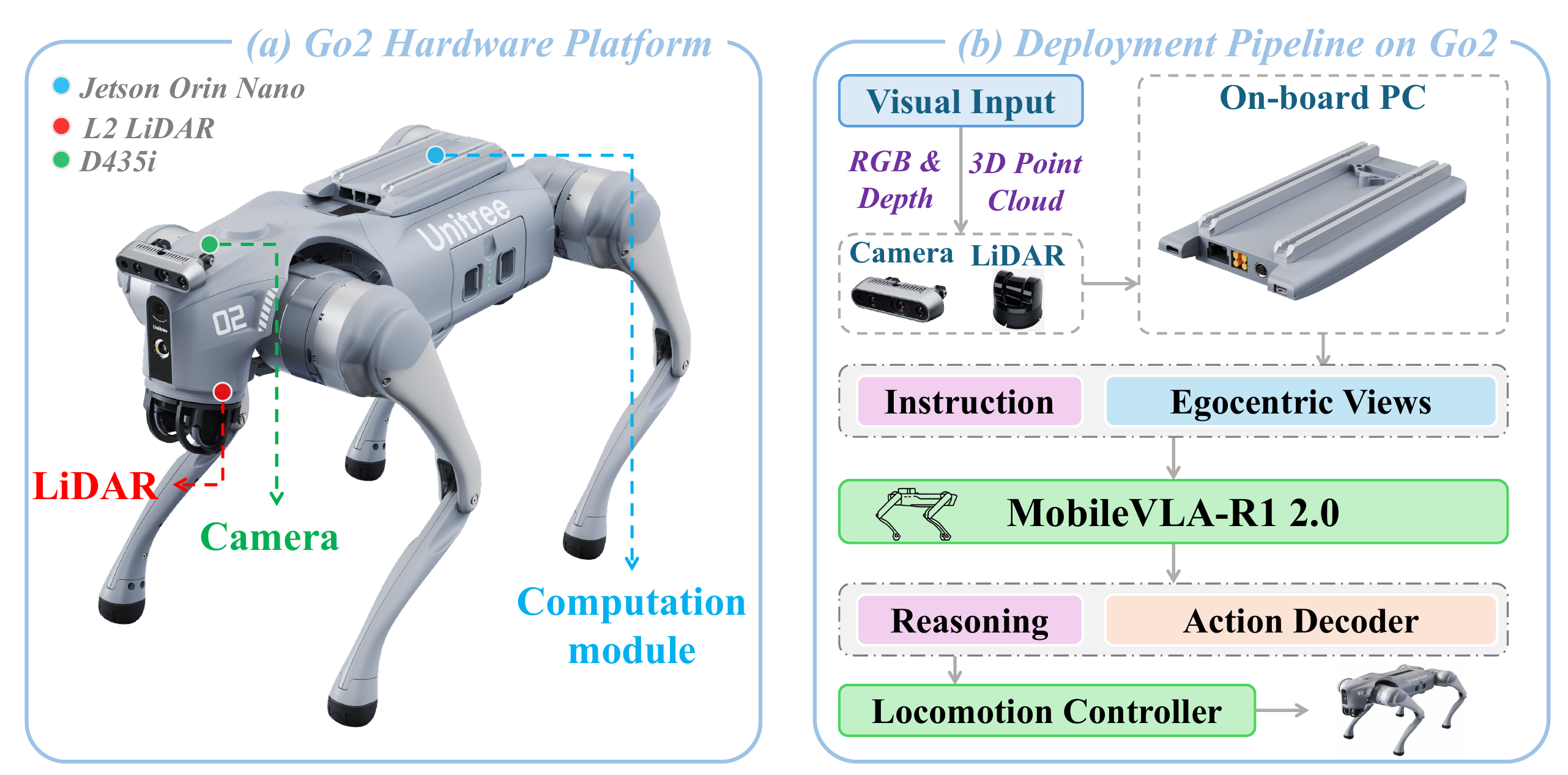}
    \caption{\textbf{Real-world hardware and deployment pipeline on Unitree Go2.}
    \textbf{(a)} Go2 platform equipped with an Intel RealSense D435i RGB-D camera, an L2 LiDAR, and a Jetson Orin Nano.
    \textbf{(b)} Hybrid Deployment pipeline. Sensing and multimodal preprocessing are performed onboard, while the 8B VLA backbone and reasoning-conditioned action decoder are executed on a remote H20 GPU.
    The resulting task-level commands $(V_x,V_y,\omega,\alpha)$ are transmitted back to the robot.
    }
    \label{fig:real_world}
    \vspace{-0.3cm}
\end{figure}

\begin{figure}[t]
    \centering
    \small
    \includegraphics[width=\linewidth]{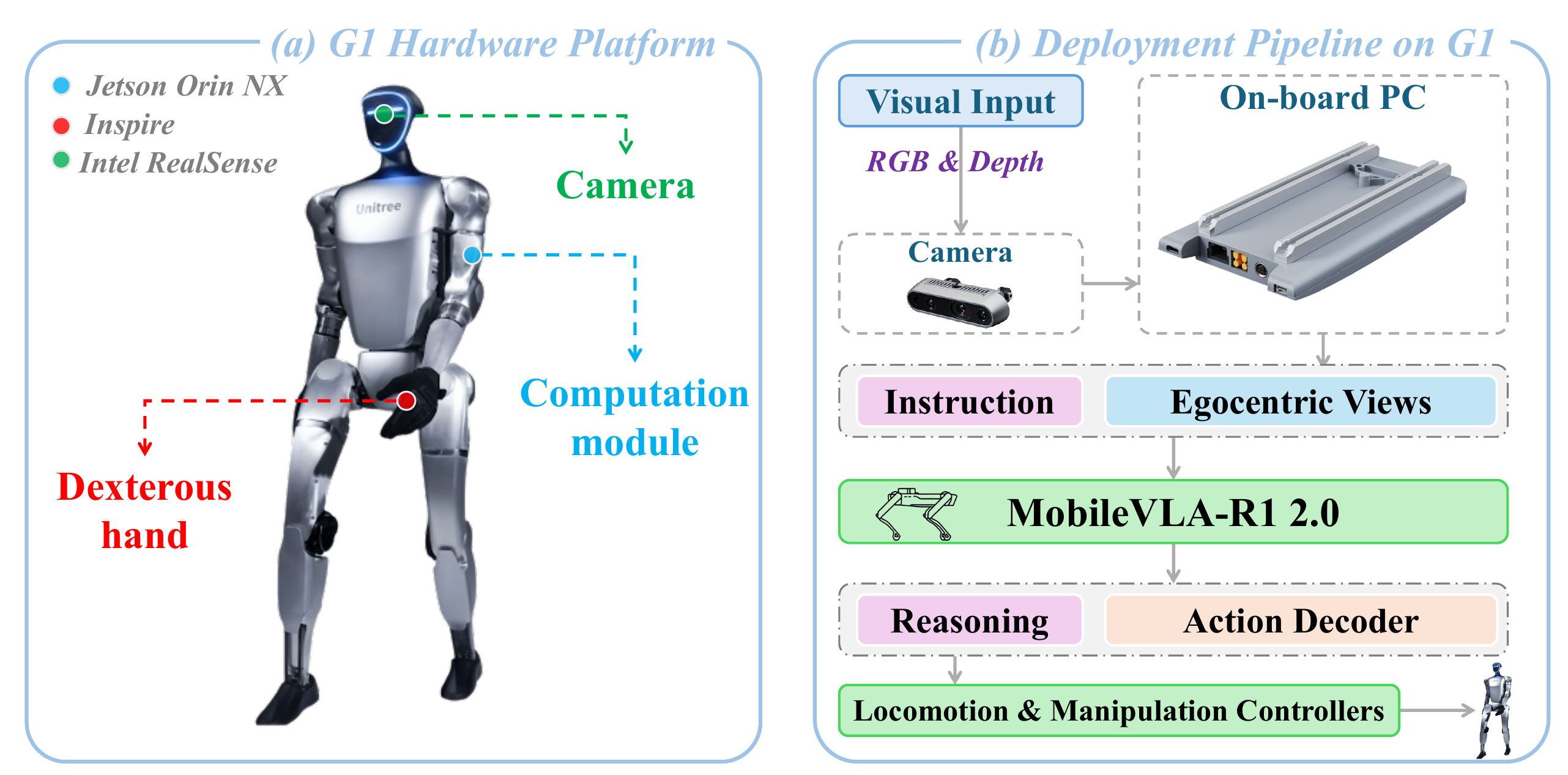}
    \caption{\textbf{Real-world hardware and deployment pipeline on Unitree G1.}
    \textbf{(a)} G1 humanoid platform equipped with an Intel RealSense RGB-D camera, an Inspire dexterous hand, and a Jetson Orin NX computation module.
    \textbf{(b)} Hybrid Deployment pipeline. RGB-D observations and language instructions are processed by \modelname{}, where the reasoning-conditioned action decoder maps multimodal representations to task-level commands.
    }
    \label{fig:real_world_g1}
    \vspace{-0.6cm}
\end{figure}
\section{Experiments}
\subsection{Experimental Setup}
\noindent\textbf{Benchmarks and metrics.}
We evaluate \modelname{} on language-guided navigation, reasoning-aligned robot control, and real-world mobile robot execution.
For navigation, we use VLN-CE~\cite{krantz2020beyond,ku2020room}, which extends R2R and RxR to continuous photorealistic environments and requires closed-loop instruction following under egocentric observations.
Following standard protocols, we report results on the \textit{val-unseen} splits of R2R-CE and RxR-CE using navigation error (NE), oracle success rate (OS), success rate (SR), success-weighted path length (SPL), and normalized dynamic time warping (nDTW).
We further evaluate task-level action generation on QUARD~\cite{ding2024quar}, which contains diverse locomotion and behavior tasks with paired observations and executable control targets.
Following the official protocol, we report task-wise success rates and the average success rate over the six tasks, with 25 evaluation episodes per task.

\begin{figure*}[!t]
    \centering
    \small
    \includegraphics[width=0.8\linewidth]{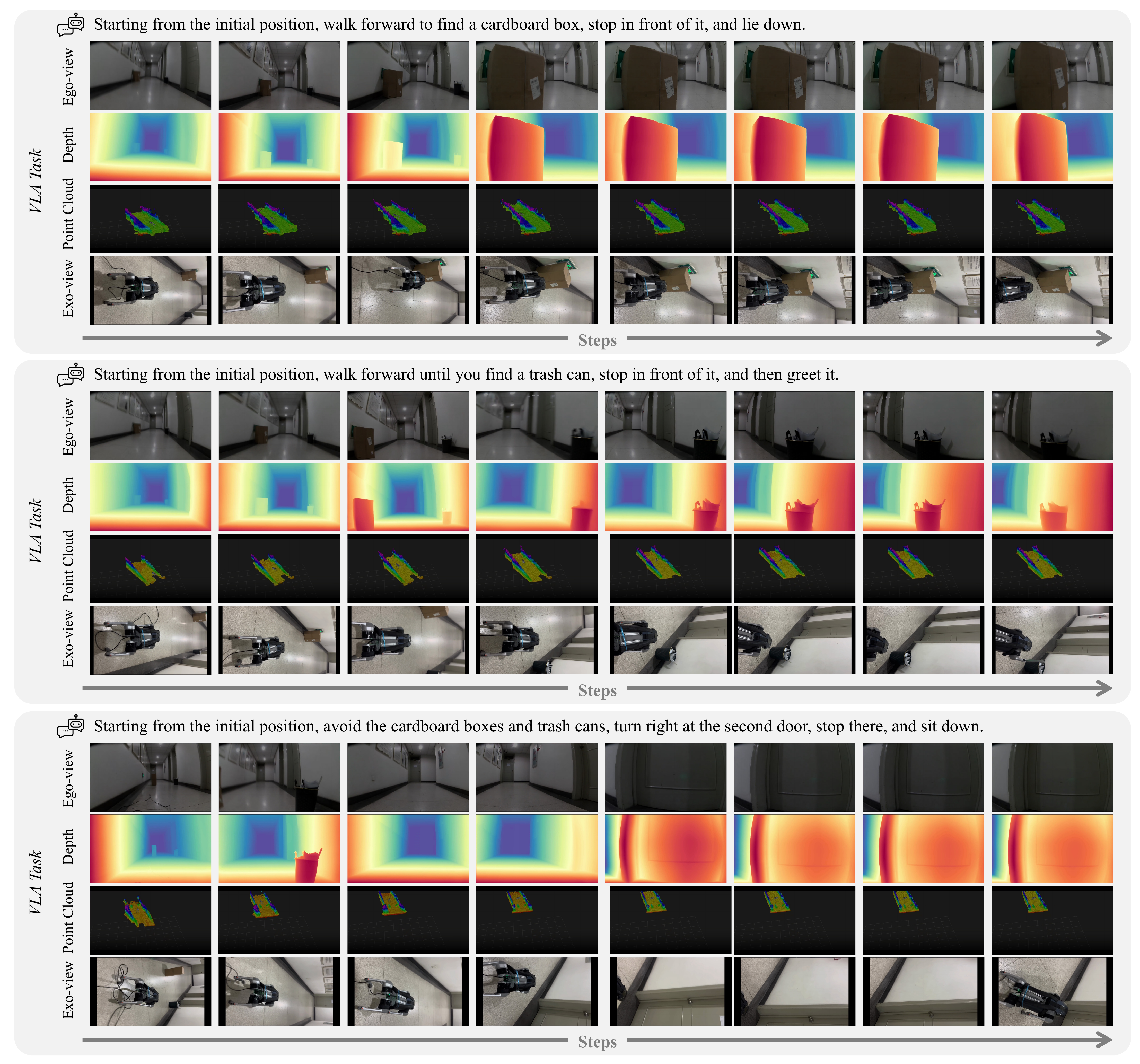}
    \caption{\textbf{Indoor qualitative results on Unitree Go2.}
    Representative closed-loop rollouts show target-directed navigation, turning, and obstacle avoidance in indoor environments.
    }
    \label{fig:realworld_vis1}
    \vspace{-0.5cm}
\end{figure*}

\begin{figure*}[!t]
    \centering
    \small
    \includegraphics[width=0.8\linewidth]{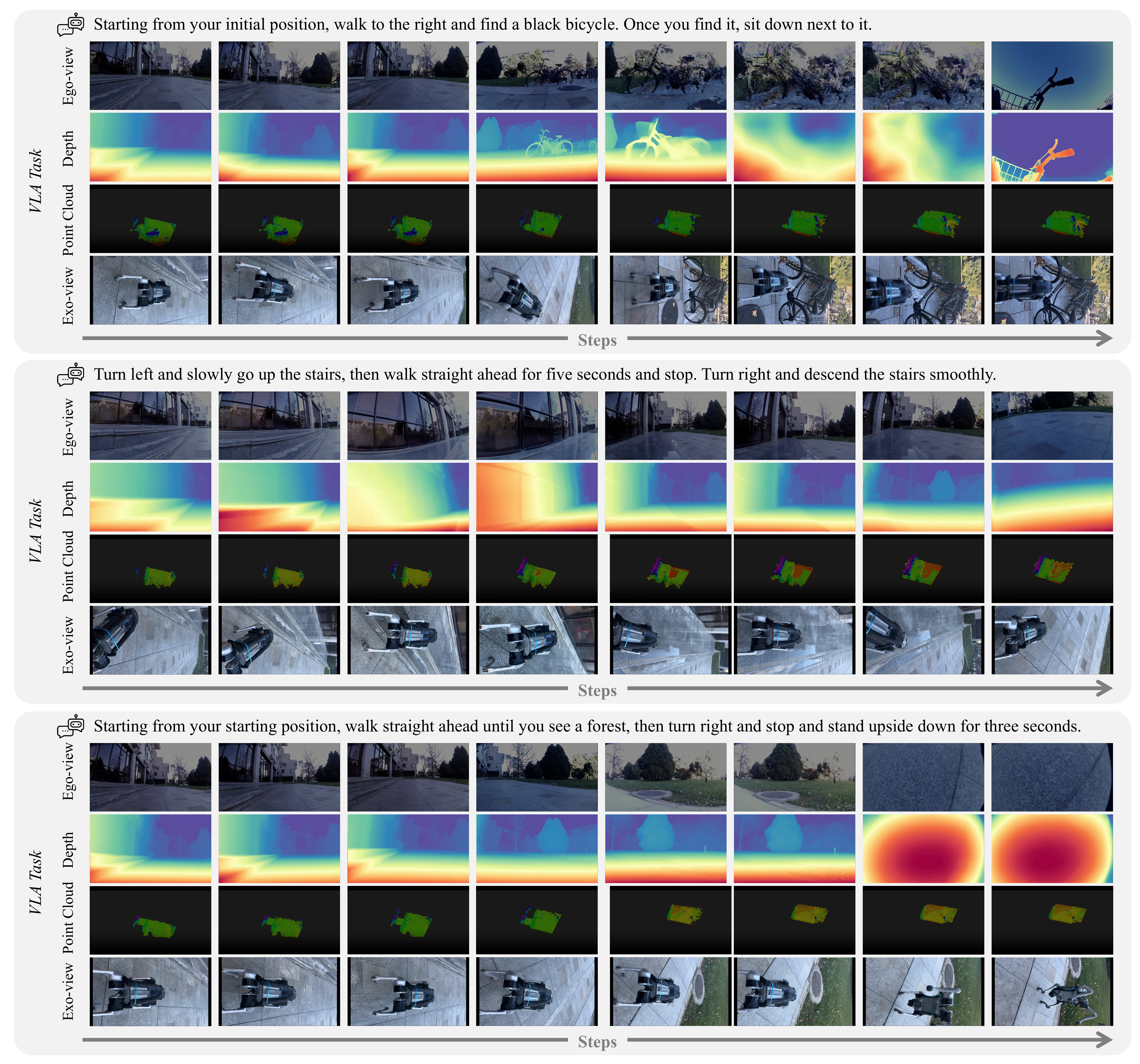}
    \caption{\textbf{Outdoor qualitative results on Unitree Go2.}
    Representative closed-loop rollouts show target-directed navigation, turning, and obstacle-aware execution in outdoor environments.
    }
    \label{fig:realworld_vis2}
    \vspace{-0.6cm}
\end{figure*}

\begin{figure*}[!t]
    \centering
    \small
    \includegraphics[width=0.8\linewidth]{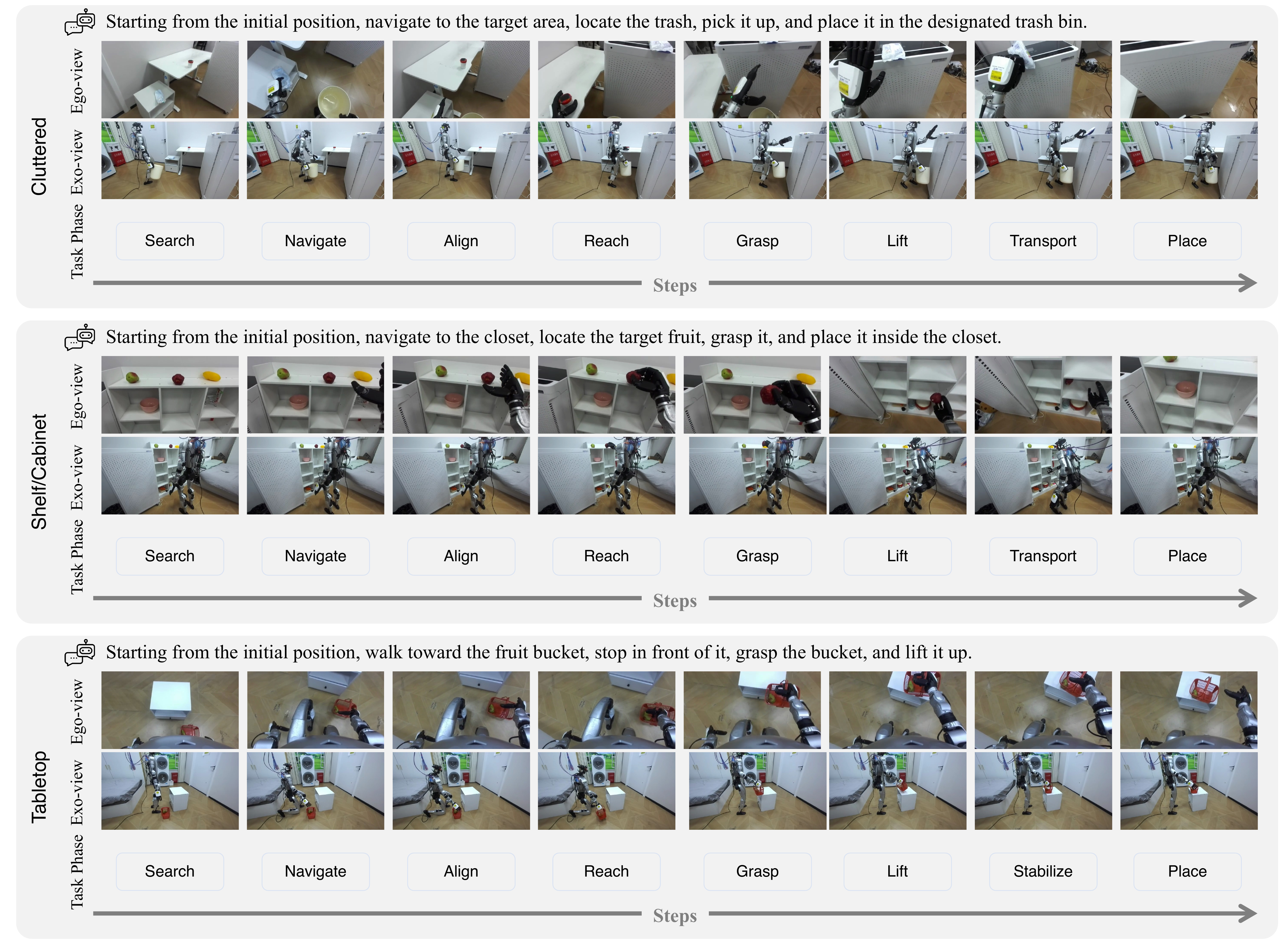}
    \caption{\textbf{Real-world qualitative results on Unitree G1.}
    Representative mobile-manipulation rollouts in the \emph{Cluttered}, \emph{Shelf/Cabinet}, and \emph{Tabletop} scenarios, showing task progression from navigation and alignment to grasping and object interaction.
    }
    \label{fig:g1_qualitative}
    \vspace{-0.6cm}
\end{figure*}

Finally, we conduct real-world evaluations on Unitree Go2 and Unitree G1.
Go2 is used to evaluate closed-loop language-guided mobile control, whereas G1 is used to evaluate humanoid mobile manipulation.
No G1-specific trajectories, demonstrations, or task annotations are used during model training.

\begin{table}[t]
\centering
\small
\caption{\textbf{Overall performance on QUARD~\cite{ding2024quar}.}
We report success rates on six tasks grouped by difficulty; Average is the mean over all tasks, with 25 episodes per task.}
\label{tab:exp_baseline}
\resizebox{\linewidth}{!}{
\begin{tabular}{lccccccc}
\toprule
\multirow{2}{*}{Method} & Easy & Medium & \multicolumn{4}{c}{Hard} & \multirow{2}{*}{Average} \\
\cmidrule(lr){2-2} \cmidrule(lr){3-3} \cmidrule(lr){4-7}
 & Distinguish & Go-to & Go-avoid & Go-through & Crawl & Unload &  \\
\midrule
CLIP~\cite{radford2021learning}     & 0.44 & 0.43 & 0.45 & 0.19 & 0.00 & 0.00 & 0.25 \\
VC-1~\cite{majumdar2023we}          & 0.46 & 0.43 & 0.45 & 0.31 & 0.00 & 0.00 & 0.28 \\
QUART~\cite{ding2024quar}           & 0.66 & 0.60 & 0.53 & 0.41 & 0.32 & 0.12 & 0.44 \\
MoRE~\cite{zhao2025more}            & 0.82 & 0.80 & 0.59 & 0.57 & 0.49 & 0.33 & 0.60 \\
\midrule
MobileVLA-R1~\cite{huang2025mobilevla}                        & 0.92 & 0.89 & 0.71 & 0.65 & 0.58 & 0.44 & 0.70 \\
\textbf{\modelname{} (Ours)}        & \textbf{0.95} & \textbf{0.92} & \textbf{0.77} & \textbf{0.72} & \textbf{0.66} & \textbf{0.56} & \textbf{0.76} \\
\bottomrule
\end{tabular}
}
\vspace{-0.6cm}
\end{table}

\noindent\textbf{Implementation details.}
\label{sec:supp_arch_details}
We initialize \modelname{} from NaVILA~\cite{cheng2025navila} with a LLaMA3-8B backbone.
DepthAnything V2~\cite{yang2024depth} and Point Transformer V3~\cite{wu2024point} are used as the depth and point-cloud encoders, respectively.
All modality encoders are kept frozen, while the multimodal projection layers and LoRA parameters of the VLA backbone are optimized during supervised alignment.

The reasoning-conditioned action decoder contains two learnable queries for locomotion and behavior prediction.
It employs a single cross-attention layer with 8 attention heads and a hidden dimension of $d=4096$, followed by lightweight prediction heads for continuous locomotion $(V_x,V_y,\omega)$ and the discrete task-level behavior primitive $\alpha$.
The decoder is jointly optimized during supervised alignment and kept fixed during GRPO.

For supervised alignment, we set $\lambda_{\omega}=0.5$, $\eta_{\mathrm{loc}}=1.0$, and $\eta_{\mathrm{beh}}=1.0$.
We use LoRA~\cite{hu2022lora} with rank $r=16$ and scaling factor $\alpha_{\mathrm{LoRA}}=32$.
SFT is performed for 3 epochs on $4\times$ H20 (96GB) GPUs using AdamW with a learning rate of $2\times10^{-4}$, a weight decay of $0.01$, a warmup ratio of $0.03$, and a cosine learning-rate schedule.

For GRPO, we sample $N=8$ candidate outputs per input and set the reward weights to $\lambda_{\mathrm{mov}}=1.0$, $\lambda_{\mathrm{beh}}=1.0$, and $\lambda_{\mathrm{fmt}}=0.2$.
The translational and angular commands are normalized using $s_v=1.0$ and $s_{\omega}=1.0$, respectively.
We optimize the reasoning policy for 1K steps on a single H20 (96GB) GPU using AdamW with a learning rate of $1\times10^{-6}$, KL coefficient $\beta=0.04$, and clipping coefficient $\epsilon_c=0.2$.
Unless otherwise specified, each update contains five input instances.

\subsection{Main Results}
\noindent\textbf{Vision-language navigation.}
Tab.~\ref{tab:vlnce} compares \modelname{} with prior VLN methods and the original MobileVLA-R1 on the R2R-CE~\cite{krantz2020beyond} and RxR-CE~\cite{ku2020room} \textit{val-unseen} splits.
On R2R-CE, \modelname{} achieves an SR of 69.8 and an SPL of 66.9, improving over MobileVLA-R1 by 1.5 and 1.7 percentage points, respectively, while reducing NE from 4.05 to 3.86.
It also outperforms the strongest prior method, CorrectNav, by 4.7 points in SR and 4.6 points in SPL.
On RxR-CE, \modelname{} obtains 73.1 SR, 68.5 SPL, and 77.6 nDTW, corresponding to improvements of 1.6, 1.7, and 1.5 points over MobileVLA-R1, respectively, while reducing NE from 3.92 to 3.71.
These consistent gains indicate that the proposed reasoning-conditioned action decoder improves the conversion of structured reasoning into executable navigation decisions.

\noindent\textbf{Reasoning-aligned robot control.}
Tab.~\ref{tab:exp_baseline} reports the results on QUARD~\cite{ding2024quar}.
\modelname{} achieves the best performance across all six tasks and improves the average success rate from 0.70 for MobileVLA-R1 to 0.76.
Compared with the strongest external baseline, MoRE~\cite{zhao2025more}, the average success rate improves from 0.60 to 0.76.
The gains over MobileVLA-R1 are relatively small on the easier Distinguish and Go-to tasks, but become more pronounced on the challenging Go-avoid, Go-through, Crawl, and Unload tasks, with absolute improvements of 0.06, 0.07, 0.08, and 0.12, respectively.
This trend suggests that explicitly conditioning task-level action prediction on intermediate reasoning representations is particularly beneficial when control requires more complex behavior selection and execution.

\noindent\textbf{Qualitative navigation results.}
Fig.~\ref{fig:simulator_vln_1} and~\ref{fig:simulator_vln_2} present representative \textit{val-unseen} episodes from R2R-CE and RxR-CE, respectively.
The examples cover unseen environments and instructions requiring multi-step spatial decisions, including target-directed navigation, turning, and obstacle-aware execution.
Across the trajectories, \modelname{} adapts its task-level decisions to the evolving visual observations and instruction context, qualitatively complementing the navigation results in Tab.~\ref{tab:vlnce}.

\subsection{Real-World Evaluation}
\label{sec:real_world}

\noindent\textbf{Closed-loop evaluation on Unitree Go2.}
We evaluate \modelname{} on a Unitree Go2 equipped with an Intel RealSense D435i RGB-D camera, an L2 LiDAR, and a Jetson Orin Nano, as shown in Fig.~\ref{fig:real_world}(a).
Synchronized RGB-D and LiDAR observations form the multimodal input.
As illustrated in Fig.~\ref{fig:real_world}(b), sensing, mapping, multimodal preprocessing, and low-level control are performed on the robot side, while the 8B VLA backbone together with the reasoning-conditioned action decoder is executed on a remote H20 GPU due to onboard memory constraints.
The remote policy returns task-level locomotion and behavior commands $(V_x,V_y,\omega,\alpha)$, which are subsequently executed by a fixed Go2 low-level controller.

We consider three environments, \texttt{Workspace}, \texttt{Corridor}, and \texttt{Outdoor}, under Simple and Complex instruction settings.
Simple tasks contain one or two short commands, whereas Complex tasks contain $3$--$5$ sequential subgoals involving longer-horizon navigation, multiple turns, and obstacle-aware execution.
Each scenario contains 5--6 tasks with five trials per task, resulting in 160 real-world episodes.
A trial is considered successful only when the instructed task is completed without human intervention.
We report success rate (SR) and navigation error (NE) in \texttt{Workspace} and \texttt{Corridor}, and SR only in \texttt{Outdoor}, where reliable global localization is unavailable.

\begin{table}[t]
\centering
\caption{\textbf{Closed-loop real-world evaluation on Unitree Go2.}
We report navigation error (NE) and success rate (SR) under Simple and Complex instructions in three representative environments.
Outdoor results use SR only because reliable global localization is unavailable.
}
\label{tab:real_go2}
\resizebox{\linewidth}{!}{
\begin{tabular}{l cccc cccc cccc}
\toprule
\multirow{3}{*}{Method} & \multicolumn{4}{c}{\texttt{Workspace}} & \multicolumn{4}{c}{\texttt{Corridor}} & \multicolumn{4}{c}{\texttt{Outdoor}} \\
\cmidrule(lr){2-5} \cmidrule(lr){6-9} \cmidrule(lr){10-13}
& \multicolumn{2}{c}{Simple} & \multicolumn{2}{c}{Complex} & \multicolumn{2}{c}{Simple} & \multicolumn{2}{c}{Complex} & \multicolumn{2}{c}{Simple} & \multicolumn{2}{c}{Complex} \\
\cmidrule(lr){2-3} \cmidrule(lr){4-5} \cmidrule(lr){6-7} \cmidrule(lr){8-9} \cmidrule(lr){10-11} \cmidrule(lr){12-13}
& NE$\downarrow$ & SR$\uparrow$ & NE$\downarrow$ & SR$\uparrow$ & NE$\downarrow$ & SR$\uparrow$ & NE$\downarrow$ & SR$\uparrow$ & NE & SR$\uparrow$ & NE & SR$\uparrow$ \\ 
\midrule
GPT-4o~\cite{hurst2024gpt}       & 2.01 & 0.67 & 2.38 & 0.33 & 1.49 & 0.53 & 3.00 & 0.00 & -- & 0.67 & -- & 0.50 \\
NaVILA~\cite{cheng2025navila}    & 1.29 & 0.83 & 1.76 & 0.80 & 1.15 & 0.89 & 1.76 & 0.67 & -- & 0.91 & -- & 0.83 \\
\midrule
MobileVLA-R1~\cite{huang2025mobilevla}                     & 1.03 & 0.93 & 1.23 & 0.91 & 0.96 & 1.00 & 1.23 & 0.86 & -- & 1.00 & -- & 0.96 \\
\textbf{\modelname{} (Ours)}     & \textbf{0.96} & \textbf{0.95} & \textbf{1.12} & \textbf{0.94} & \textbf{0.90} & \textbf{1.00} & \textbf{1.11} & \textbf{0.91} & -- & \textbf{1.00} & -- & \textbf{0.98} \\
\bottomrule
\end{tabular}
}
\vspace{-0.4cm}
\end{table}
Tab.~\ref{tab:real_go2} summarizes the quantitative results.
\modelname{} consistently improves over MobileVLA-R1 under the same sensing and deployment protocol.
The gains are relatively small on Simple tasks, where the previous model already approaches saturation, but become more pronounced under Complex instructions.
For example, SR increases from 0.91 to 0.94 in \texttt{Workspace} and from 0.86 to 0.91 in \texttt{Corridor}, while NE decreases from 1.23 to 1.12 and from 1.23 to 1.11, respectively.
These results suggest that conditioning task-level action prediction on intermediate reasoning representations is particularly beneficial for longer-horizon physical execution.

\noindent\textbf{Go2 deployment efficiency and failure analysis.}
Tab.~\ref{tab:real_protocol} further reports end-to-end latency and episode-level failure statistics.
Latency is measured from observation acquisition to task-level command handoff and includes multimodal preprocessing, serialization, network communication, remote VLA inference, action decoding, command transmission, and control handoff.
Across the 160 episodes, the observed failures mainly involve target grounding, perception and obstacle handling, excessive turning, narrow-passage navigation, and localization drift.
Representative cases include incorrect target or intermediate-subgoal selection under distractors, inaccurate motion decisions in constrained passages, and accumulated localization drift in outdoor environments.
These observations identify perception, spatial grounding, localization, and constrained motion execution as important remaining challenges in real-world deployment.

\begin{table}[t]
\centering
\small
\caption{
\textbf{Deployment statistics and failure analysis on Unitree Go2.}
Latency is reported as the mean$\pm$standard deviation of the hybrid end-to-end latency measured from observation acquisition to task-level command handoff, and failures are counted at the episode level.
``Target,'' ``Percep.,'' ``Turn,'' ``Pass.,'' and ``Drift'' denote target-grounding, perception/obstacle, excessive-turning, narrow-passage, and localization-drift failures, respectively.
}
\label{tab:real_protocol}
\resizebox{\linewidth}{!}{
\begin{tabular}{lccccc}
\toprule
Setting & Tasks & Trials & Episodes & Latency (ms) & Failure breakdown \\
\midrule
Workspace-Simple  & 6 & 5 & 30 & $210\pm14$ & Target 1 / Percep. 1 \\
Workspace-Complex & 6 & 5 & 30 & $218\pm16$ & Target 2 / Percep. 1 \\
Corridor-Simple   & 5 & 5 & 25 & $205\pm13$ & – \\
Corridor-Complex  & 5 & 5 & 25 & $216\pm17$ & Turn 2 / Pass. 2 \\
Outdoor-Simple    & 5 & 5 & 25 & $232\pm21$ & – \\
Outdoor-Complex   & 5 & 5 & 25 & $245\pm24$ & Drift 1 \\
\midrule
Total             & 32 & – & 160 & – & 10 failures \\
\bottomrule
\end{tabular}}
\vspace{-0.6cm}
\end{table}

\noindent\textbf{Mobile manipulation on Unitree G1.}
We further evaluate \modelname{} on a Unitree G1 humanoid platform equipped with an Intel RealSense RGB-D camera, an Inspire dexterous hand, and a Jetson Orin NX computation module, as shown in Fig.~\ref{fig:real_world_g1}(a).
The deployment pipeline is illustrated in Fig.~\ref{fig:real_world_g1}(b).
RGB-D observations and language instructions are processed by \modelname{}, whose reasoning-conditioned action decoder generates task-level commands $(V_x,V_y,\omega,\alpha)$ that are executed through fixed locomotion and manipulation controllers.
No G1-specific trajectories, demonstrations, task annotations, or policy fine-tuning are used, and the learned policy remains fixed during deployment.
The same task-level action representation is retained across embodiments.
The continuous components $(V_x,V_y,\omega)$ are executed by the fixed G1 locomotion controller, while $\alpha$ selects predefined manipulation primitives.
Therefore, manipulation behaviors, including reaching, grasping, lifting, transporting, and placing, are realized through embodiment-specific controllers rather than direct joint-level prediction.

We consider three scenarios with increasing execution complexity: \texttt{Tabletop}, \texttt{Shelf/Cabinet}, and \texttt{Cluttered}.
The \texttt{Tabletop} setting emphasizes target approach and near-field manipulation; \texttt{Shelf/Cabinet} additionally requires accurate manipulation-site alignment; and \texttt{Cluttered} combines obstacle-aware navigation, target approach, and manipulation under tighter spatial constraints.
We report navigation success (Nav.), manipulation success (Manip.), and full-task success (Full).
Manip. is computed over episodes with successful navigation, whereas Full requires both navigation and manipulation to succeed within the same closed-loop episode without human intervention.

\begin{table}[t]
\centering
\small
\caption{
\textbf{Real-world mobile manipulation on Unitree G1.}
Navigation (Nav.), manipulation (Manip.), and full-task (Full) success rates are reported in percentage.
Manip. denotes the manipulation success rate conditioned on successful navigation, whereas Full requires both navigation and manipulation to succeed within the same episode.
All methods use the same sensing and low-level control interfaces, with no G1-specific policy fine-tuning.
}
\label{tab:g1_real}
\resizebox{\linewidth}{!}{
\begin{tabular}{lcccccccccc}
\toprule
\multirow{2}{*}{Method} & \multicolumn{3}{c}{\texttt{Tabletop}} & \multicolumn{3}{c}{\texttt{Shelf/Cabinet}} & \multicolumn{3}{c}{\texttt{Cluttered}} & \multirow{2}{*}{Full Avg.} \\
\cmidrule(lr){2-4} \cmidrule(lr){5-7} \cmidrule(lr){8-10}
& Nav. & Manip. & Full & Nav. & Manip. & Full & Nav. & Manip. & Full & \\
\midrule
NaVILA~\cite{cheng2025navila} & 75.0 & 63.3 & 47.5 & 67.5 & 59.3 & 40.0 & 57.5 & 43.5 & 25.0 & 37.5 \\

MobileVLA-R1~\cite{huang2025mobilevla}                  & 80.0 & 71.9 & 57.5 & 72.5 & 65.5 & 47.5 & 65.0 & 53.8 & 35.0 & 46.7 \\ \midrule
\textbf{\modelname{} (Ours)}  & \textbf{85.0} & \textbf{79.4} & \textbf{67.5} & \textbf{77.5} & \textbf{74.2} & \textbf{57.5} & \textbf{70.0} & \textbf{64.3} & \textbf{45.0} & \textbf{56.7} \\
\bottomrule
\end{tabular}
}
\vspace{-0.6cm}
\end{table}
Tab.~\ref{tab:g1_real} summarizes the G1 results.
\modelname{} consistently outperforms NaVILA and MobileVLA-R1 across all three scenarios.
Compared with MobileVLA-R1, the average full-task success rate increases from 46.7\% to 56.7\%, corresponding to an absolute improvement of 10.0 percentage points.
Conditional manipulation success increases from 71.9\% to 79.4\% on \texttt{Tabletop}, from 65.5\% to 74.2\% on \texttt{Shelf/Cabinet}, and from 53.8\% to 64.3\% on \texttt{Cluttered}.
The consistent improvements in conditional manipulation success and full-task completion support the effectiveness of reasoning-conditioned action prediction for translating structured reasoning into executable mobile-manipulation decisions.
Importantly, these gains are obtained without G1-specific policy fine-tuning, supporting evaluation-only transfer of the learned task-level reasoning-to-action interface to a humanoid platform.

\noindent\textbf{G1 evaluation scale and failure analysis.}
Tab.~\ref{tab:g1_protocol} summarizes the evaluation scale and dominant failure categories across the three G1 scenarios.
Failures are assigned to their primary observed cause, including target-grounding, navigation/positioning, grasping, manipulation-execution, and low-level control errors.
Across the 120 episodes, grasping and manipulation-execution errors are the most frequent failure categories, with 16 and 14 cases, respectively, followed by navigation/positioning errors (10 cases), target-grounding errors (8 cases), and low-level control failures (4 cases).
The \texttt{Cluttered} setting exhibits more grounding, positioning, and interaction failures than the simpler settings, indicating that coordinating perception, mobility, and manipulation becomes increasingly challenging as geometric and perceptual complexity increases.
\begin{table}[t]
\centering
\small
\caption{
\textbf{Evaluation scale and failure statistics on Unitree G1.}
Failures are counted at the episode level and categorized according to their dominant observed cause.
``Ground.,'' ``Nav.,'' ``Grasp,'' ``Manip.,'' and ``Ctrl.'' denote target-grounding, navigation/positioning, grasping, manipulation-execution, and low-level control failures, respectively.
}
\label{tab:g1_protocol}
\resizebox{\linewidth}{!}{
\begin{tabular}{lccccl}
\toprule
Scenario & Tasks & Trials/Task & Episodes & Successful & Failure Breakdown \\
\midrule
Tabletop & 4 & 10 & 40 & 27 & Ground. 2 / Nav. 2 / Grasp 5 / Manip. 3 / Ctrl. 1 \\
Shelf/Cabinet & 4 & 10 & 40 & 23 & Ground. 2 / Nav. 3 / Grasp 5 / Manip. 6 / Ctrl. 1 \\
Cluttered & 4 & 10 & 40 & 18 & Ground. 4 / Nav. 5 / Grasp 6 / Manip. 5 / Ctrl. 2 \\
\midrule
Total & 12 & -- & 120 & 68 & Ground. 8 / Nav. 10 / Grasp 16 / Manip. 14 / Ctrl. 4 \\
\bottomrule
\end{tabular}
}
\vspace{-0.6cm}
\end{table}

\noindent\textbf{Qualitative real-world results.}
Fig.~\ref{fig:realworld_vis1} and Fig.~\ref{fig:realworld_vis2} present representative indoor and outdoor rollouts on Unitree Go2, while Fig.~\ref{fig:g1_qualitative} shows mobile-manipulation rollouts on Unitree G1 across the \emph{Tabletop}, \emph{Shelf/Cabinet}, and \emph{Cluttered} scenarios.

\subsection{Ablation Study}
\noindent\textbf{Effect of action decoding and conditioning.}
Tab.~\ref{tab:decoder_ablation} evaluates the proposed action decoder under different conditioning inputs.
Replacing deterministic parsing with an observation-conditioned learnable decoder consistently improves performance, while conditioning on intermediate reasoning representations yields larger gains.
Combining observation and reasoning representations achieves the best performance, increasing SR from 68.3 to 69.8 and SPL from 65.2 to 66.9 on R2R-CE.
On RxR-CE, SR improves from 71.5 to 73.1, while SPL increases from 66.8 to 68.5.
These results indicate that intermediate reasoning states provide action-relevant information complementary to perceptual observations, supporting their joint use for task-level action prediction.

\begin{table}[bhtp]
\vspace{-0.3cm}
\centering
\small
\caption{
\textbf{Ablation of action decoding and conditioning.}
We compare deterministic action parsing with learnable action decoders under different conditioning inputs.
``Obs.'' and ``Reason.'' denote observation and intermediate reasoning representations, respectively.
}
\label{tab:decoder_ablation}
\resizebox{\linewidth}{!}{
\begin{tabular}{lcc|cccc}
\toprule
\multirow{2}{*}{Action Interface} & \multirow{2}{*}{Obs.} & \multirow{2}{*}{Reason.} & \multicolumn{2}{c}{R2R-CE} & \multicolumn{2}{c}{RxR-CE} \\
\cmidrule(lr){4-5} \cmidrule(lr){6-7}
& & & SR$\uparrow$ & SPL$\uparrow$ & SR$\uparrow$ & SPL$\uparrow$ \\
\midrule
Deterministic Parsing & \xmark & \xmark & 68.3 & 65.2 & 71.5 & 66.8 \\
Learnable Decoder     & \cmark & \xmark & 68.8 & 65.7 & 72.0 & 67.3 \\
Learnable Decoder     & \xmark & \cmark & 69.3 & 66.3 & 72.6 & 67.9 \\ \midrule
\textbf{Learnable Decoder} & \cmark & \cmark & \textbf{69.8} & \textbf{66.9} & \textbf{73.1} & \textbf{68.5} \\
\bottomrule
\end{tabular}
}
\vspace{-0.3cm}
\end{table}

\noindent\textbf{Effect of the action decoder and GRPO optimization.}
Tab.~\ref{tab:component_ablation} evaluates the individual and joint effects of the proposed action decoder and GRPO optimization.
Either component consistently improves performance over the supervised baseline, with GRPO providing the larger individual gain.
Their combination performs best, increasing SR from 68.3 to 69.8 and SPL from 65.2 to 66.9 over the GRPO-only variant on R2R-CE, with consistent improvements on RxR-CE.
These results show that learnable action decoding and GRPO optimization provide complementary benefits for reasoning-to-action alignment.

\begin{table}[bhtp]
\vspace{-0.4cm}
\centering
\small
\caption{
\textbf{Ablation of the action decoder and GRPO optimization.}
We evaluate the individual and joint effects of the two components.
Without the action decoder, task-level actions are obtained by deterministic parsing; without GRPO, the model is trained with supervised learning only.
}
\label{tab:component_ablation}
\resizebox{0.85\linewidth}{!}{
\begin{tabular}{cc|cccc}
\toprule
\multirow{2}{*}{Action Decoder} & \multirow{2}{*}{GRPO} & \multicolumn{2}{c}{R2R-CE} & \multicolumn{2}{c}{RxR-CE} \\
\cmidrule(lr){3-4} \cmidrule(lr){5-6}
& & SR$\uparrow$ & SPL$\uparrow$ & SR$\uparrow$ & SPL$\uparrow$ \\
\midrule
\xmark & \xmark & 58.0 & 53.2 & 61.4 & 56.7 \\
\cmark & \xmark & 63.1 & 58.7 & 66.0 & 61.4 \\
\xmark & \cmark & 68.3 & 65.2 & 71.5 & 66.8 \\
\midrule
\cmark & \cmark & \textbf{69.8} & \textbf{66.9} & \textbf{73.1} & \textbf{68.5} \\
\bottomrule
\end{tabular}
}
\vspace{-0.4cm}
\end{table}

\noindent\textbf{Effect of reasoning supervision granularity.}
Tab.~\ref{tab:cot_necessity} evaluates different granularities of reasoning supervision under the same training configuration.
All single-granularity variants outperform No-CoT, with navigation-level supervision providing the strongest individual performance.
Combining episode-, navigation-, and step-level supervision achieves the best results, improving SR from 64.0 to 68.3 and SPL from 59.6 to 65.2.
This indicates that reasoning signals at different temporal granularities provide complementary supervision for long-horizon instruction following.

\begin{table}[t]
\centering
\small
\caption{
\textbf{Ablation of reasoning supervision granularity on R2R-CE Val-Unseen.}
We vary the granularity of reasoning supervision while keeping the teacher model, action targets, output format, and training budget fixed.
No-CoT removes rationale supervision while preserving identical action supervision.
}
\label{tab:cot_necessity}
\resizebox{0.8\linewidth}{!}{
\begin{tabular}{lcc}
\toprule
Reasoning Granularity & SR$\uparrow$ & SPL$\uparrow$ \\
\midrule
No-CoT          & 64.0 & 59.6 \\
Episode-level   & 65.5 & 61.3 \\
Step-level      & 66.0 & 61.7 \\
Navigation-level& 66.7 & 62.6 \\
\midrule
\textbf{Multi-granularity} & \textbf{68.3} & \textbf{65.2} \\
\bottomrule
\end{tabular}
}
\vspace{-0.6cm}
\end{table}

\noindent\textbf{Effect of GRPO reward components.}
Tab.~\ref{tab:ablation_rewards} evaluates the GRPO reward design under the deterministic action interface.
All individual reward terms improve over the supervised baseline, with the behavior reward yielding the strongest single-component performance.
Combining reward terms consistently provides further gains, and the full configuration achieves the best SR of 68.3 and SPL of 65.2.
This indicates that movement alignment, behavior correctness, and output-format validity provide complementary signals for GRPO optimization.
\begin{table}[bhtp]
\vspace{-0.3cm}
\centering
\small
\caption{
\textbf{Ablation of GRPO reward components on R2R-CE Val-Unseen.}
Results are obtained with deterministic action parsing.
We evaluate the movement, behavior, and format rewards; the configuration without reward optimization denotes the supervised pre-GRPO baseline.
}
\label{tab:ablation_rewards}
\resizebox{0.7\linewidth}{!}{
\begin{tabular}{ccc|cc}
\toprule
$R_{\rm mov}$ & $R_{\rm beh}$ & $R_{\rm fmt}$
& SR$\uparrow$ & SPL$\uparrow$ \\
\midrule
\xmark & \xmark & \xmark & 58.0 & 53.2 \\
\midrule
\cmark & \xmark & \xmark & 60.7 & 55.5 \\
\xmark & \cmark & \xmark & 61.9 & 56.8 \\
\xmark & \xmark & \cmark & 59.6 & 54.7 \\
\midrule
\cmark & \cmark & \xmark & 64.5 & 60.2 \\
\cmark & \xmark & \cmark & 63.4 & 59.1 \\
\xmark & \cmark & \cmark & 65.2 & 61.0 \\
\midrule
\cmark & \cmark & \cmark & \textbf{68.3} & \textbf{65.2} \\
\bottomrule
\end{tabular}
}
\vspace{-0.5cm}
\end{table}

\noindent\textbf{Effect of action decoder architecture.}
Tab.~\ref{tab:decoder_design} evaluates different decoder architectures under the same representations and training protocol.
Attention-based decoding consistently outperforms mean pooling, while separating the locomotion and behavior branches provides further gains.
The proposed Dual-Query Dual-Head decoder achieves the best performance, reaching 69.8 SR and 66.9 SPL, corresponding to improvements of 1.1 and 1.3 points over Mean Pooling + MLP, respectively.
This indicates that dedicated queries for locomotion and behavior prediction provide a more effective mechanism for extracting action-relevant information from the joint observation and reasoning representations.

\begin{table}[bhtp]
\centering
\small
\caption{
\textbf{Ablation of action decoder architectures on R2R-CE Val-Unseen.}
All variants use identical observation and reasoning representations and the same training protocol, differing only in the decoder architecture.
}
\label{tab:decoder_design}
\resizebox{0.8\linewidth}{!}{
\begin{tabular}{lcc}
\toprule
Action Decoder & SR$\uparrow$ & SPL$\uparrow$ \\
\midrule
Mean Pooling + MLP       & 68.7 & 65.6 \\
Single-Query Attention   & 69.1 & 66.1 \\
Shared-Query Dual-Head   & 69.4 & 66.5 \\
\midrule
\textbf{Dual-Query Dual-Head}  & \textbf{69.8} & \textbf{66.9} \\
\bottomrule
\end{tabular}
}
\vspace{-0.3cm}
\end{table}

Additional analyzes of multimodal perception, reward-weight sensitivity, rationale sources, and optimization objectives are provided in \textit{App.}~A.

\section{Test-Time Efficiency}
Efficient inference is essential for deploying large VLA policies on resource-constrained mobile robots.
Our real-world system adopts a hybrid deployment architecture: sensing, mapping, multimodal preprocessing, and low-level control are performed on the robot side, while the 8B VLA backbone and the reasoning-conditioned action decoder are executed on a remote H20 GPU due to the memory constraints of the onboard Jetson Orin Nano.
The remote policy directly produces task-level commands $(V_x,V_y,\omega,\alpha)$, which are transmitted back to the robot for execution without deterministic parsing of textual action outputs.

Measured from observation acquisition to task-level command handoff, the end-to-end latency ranges from 205 to 245\,ms across real-world scenarios.
This measurement includes multimodal preprocessing, serialization, network communication, remote VLA inference, action decoding, command transmission, and control handoff, corresponding to an effective high-level decision rate of approximately 4.1--4.9\,Hz.
The robot-specific low-level controllers operate independently at higher control frequencies.

The 8B VLA backbone remains the primary computational and memory bottleneck, preventing fully onboard inference on the current Jetson platform.
Remote inference may therefore introduce additional latency and reduced robustness under constrained or unstable network connectivity.
Model compression, quantization, distillation, and lightweight VLA backbones are promising directions toward fully onboard deployment.

\section{Limitations and Future Work}

\noindent\textbf{Limitations.}
Despite the improvements of \modelname{}, several limitations remain.
\emph{(i) Task-level action abstraction.}
Although the reasoning-conditioned action decoder provides a learnable interface between reasoning and physical execution, the current action space remains task-level, combining continuous locomotion commands with a finite set of behavior primitives.
It therefore does not directly represent fine-grained contact dynamics, dexterous manipulation, or continuously parameterized whole-body motion.
\emph{(ii) Training and embodiment coverage.}
The current model is trained using supervised trajectories and multi-granularity reasoning annotations from the considered navigation and robot-control datasets.
While the G1 experiments demonstrate transfer without G1-specific policy fine-tuning, they do not address general embodiment adaptation.
Performance may therefore degrade under substantially different robot morphologies, sensing configurations, environments, or task distributions.
\emph{(iii) Hybrid deployment.}
The current real-world system executes the 8B VLA backbone on a remote GPU because of onboard memory constraints.
Although the resulting latency supports the closed-loop tasks evaluated in this work, remote inference introduces communication overhead and dependence on network connectivity.

\noindent\textbf{Future work.}
Future work will investigate richer parameterized skills and hierarchical whole-body control to improve action expressiveness, broader cross-platform training and online adaptation to improve embodiment and environment coverage, and model compression, quantization, and distillation toward fully onboard inference.
Extending the framework to dexterous manipulation, dynamic human--robot interaction, and longer-horizon open-world tasks also represents an important direction.

\section{Conclusion}
In this work, we present \modelname{}, a reasoning-enhanced VLA framework for language-guided mobile robot control.
By combining multi-granularity reasoning supervision, a reasoning-conditioned action decoder, and GRPO-based optimization, \modelname{} directly connects intermediate reasoning representations with task-level locomotion and behavior prediction.
Experiments on R2R-CE, RxR-CE, and QUARD demonstrate consistent improvements in navigation and reasoning-aligned control.
Real-world evaluations on Unitree Go2 and Unitree G1 further validate closed-loop mobile control and humanoid mobile manipulation, respectively, with the G1 evaluation requiring no embodiment-specific policy fine-tuning.
These results highlight the value of explicit reasoning-to-action alignment for developing interpretable and executable VLA policies for long-horizon embodied tasks.

\ifCLASSOPTIONcompsoc
 \section*{Acknowledgments}
 This work is supported by the Fundamental Research Funds for the Central Universities, Peking University.
\else
 \section*{Acknowledgment}
\fi

{
\bibliographystyle{IEEEtran}
\bibliography{reference}
}

\begin{IEEEbiography}[{\includegraphics[width=1in]{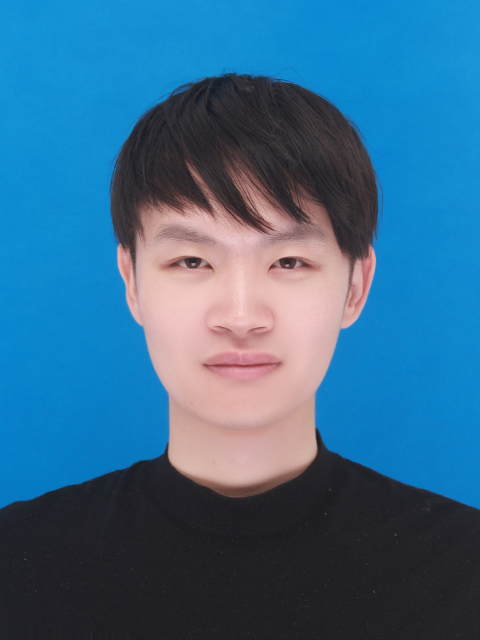}}]{Ting Huang}
is a researcher in embodied intelligence and multimodal AI, advised by Prof. Hao Tang. He received his master’s degree in Control Science and Engineering from Shanghai University of Engineering Science. His research interests lie in 3D spatial intelligence, multimodal foundation models, and embodied AI, aiming at unified learning frameworks that integrate geometric scene understanding, spatial reasoning, and vision-language-action learning for agents operating in complex physical environments.
\end{IEEEbiography}

\begin{IEEEbiography}[{\includegraphics[width=1in]{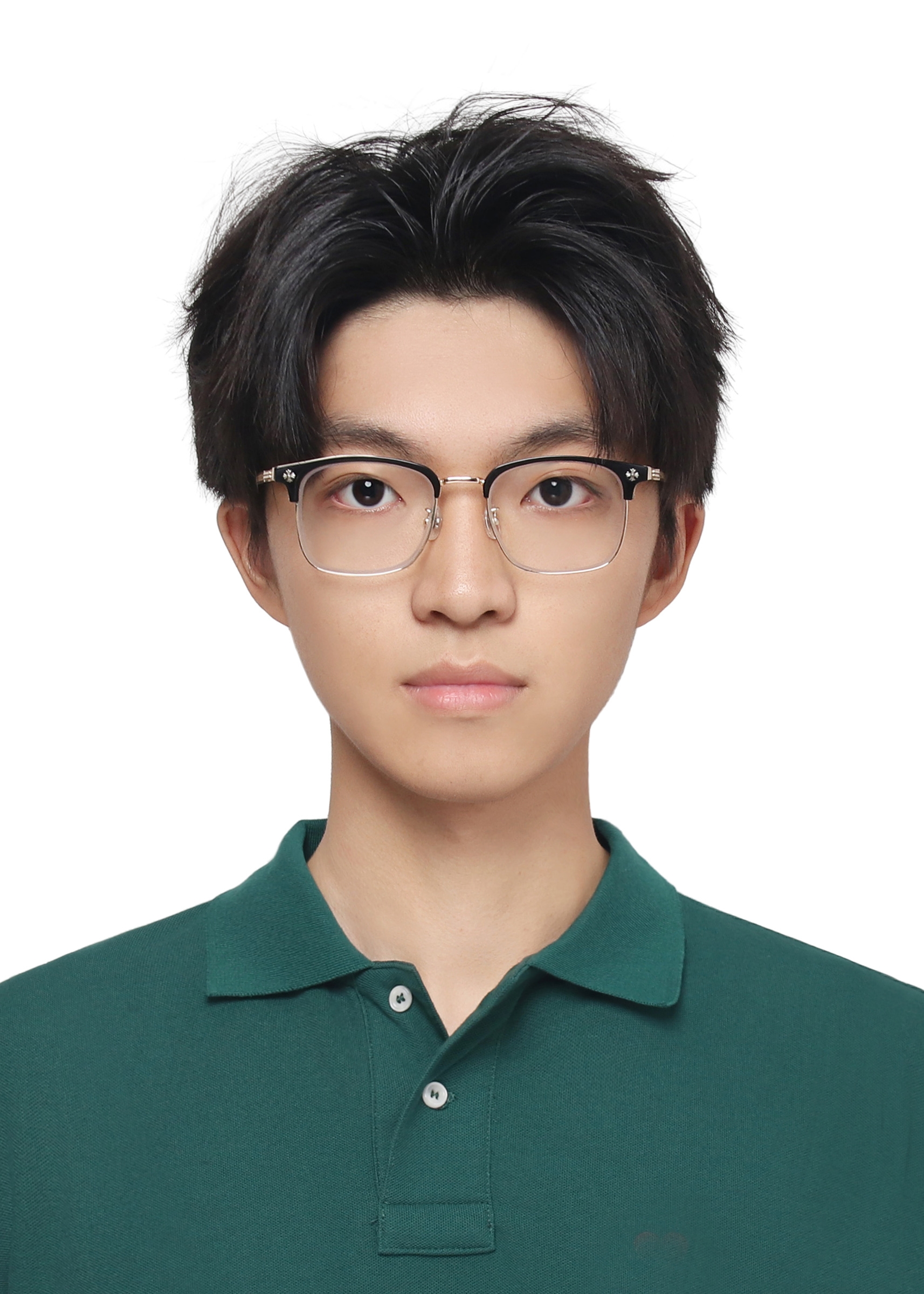}}]{Yue Huang}
 is an undergraduate student at South China University of Technology. Currently, his main research interests lie in Humanoid locomotion, 3D perspective modeling, world models, and quantitative investment.
\end{IEEEbiography}

\begin{IEEEbiography}[{\includegraphics[width=1in]{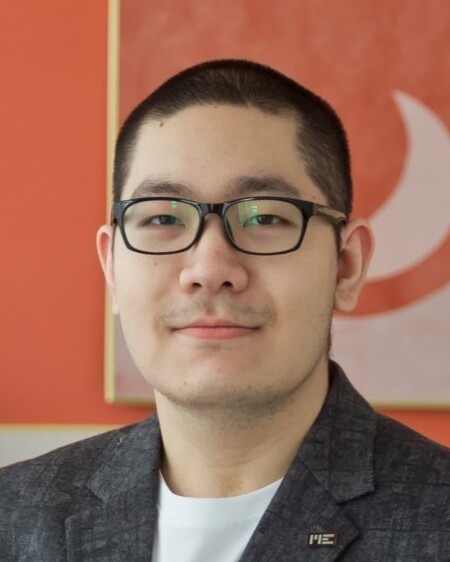}}]{Zeyu Zhang}
is a researcher working on generative AI, with a particular interest in building models that understand and interact with the physical world. He received his bachelor’s degree from the Australian National University, where he was advised by Prof. Richard Hartley and Prof. Ian Reid. His research explores generative modeling for learning physical dynamics from visual data. His work spans world models, multimodal foundation models, embodied AI, and AI for health. 
\end{IEEEbiography}

\begin{IEEEbiography}[{\includegraphics[width=1in]{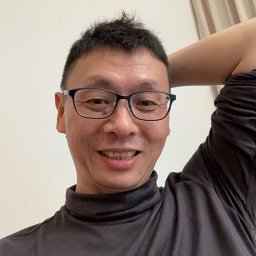}}]{Shuicheng Yan}
is a Distinguished Professor (Practice) at the National University of Singapore (NUS), Singapore. Previously, he served as the Group Chief Scientist at Sea Group and held several senior research and industry positions. He received his B.S. and Ph.D. degrees from Peking University, China.
He is a Fellow of the Singapore Academy of Engineering, AAAI, ACM, IEEE, and IAPR.
His research interests include computer vision, multimedia analysis, and efficient artificial general intelligence.
\end{IEEEbiography}

\begin{IEEEbiography}[{\includegraphics[width=1in,height=1.25in,clip,keepaspectratio]{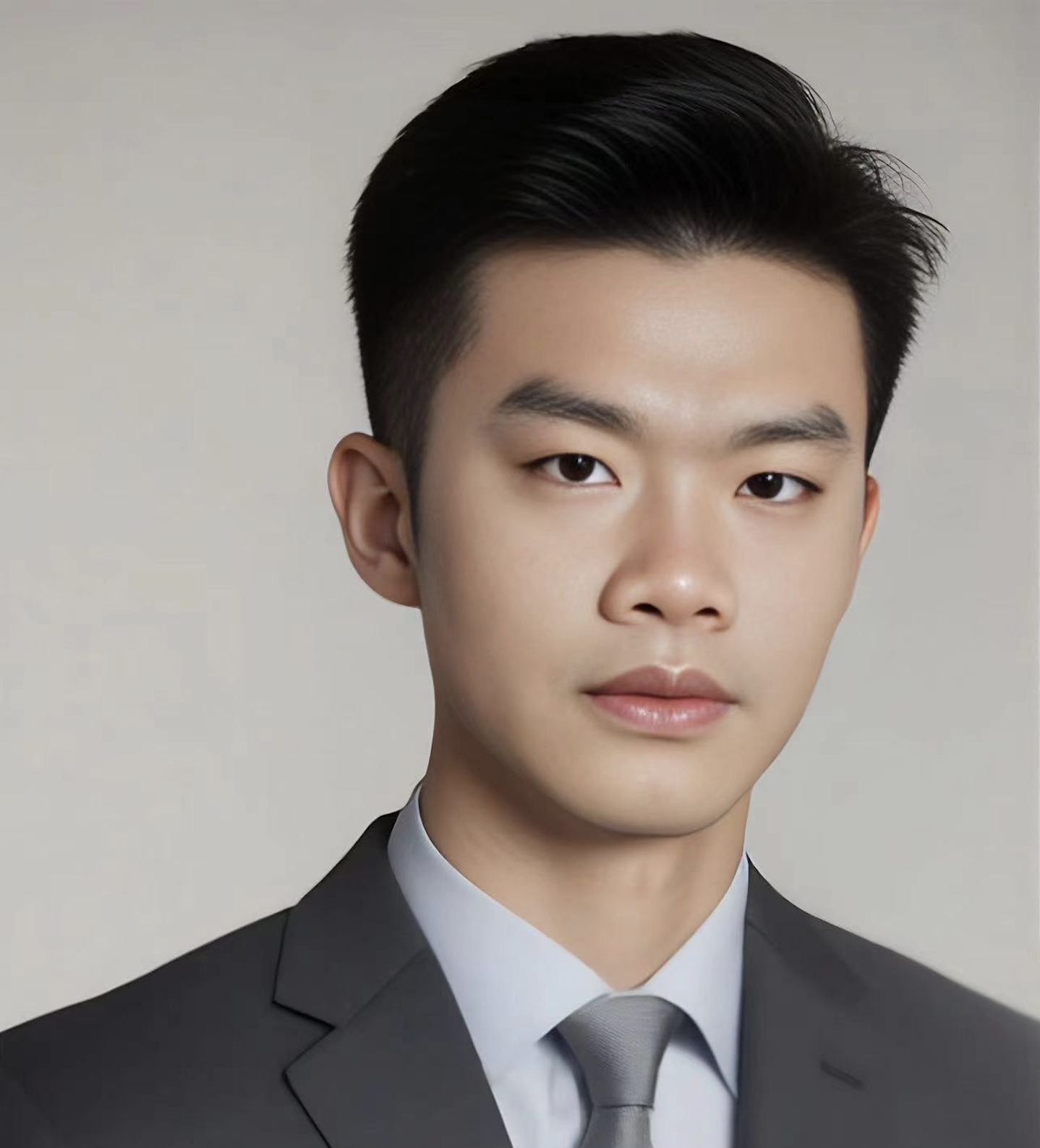}}]{Hao Tang}
is an Assistant Professor at Peking University, China. Previously, he held postdoctoral positions at CMU, USA, and ETH Zürich, Switzerland. He earned his master’s degree from Peking University, and his Ph.D. from the University of Trento, Italy.
He has had the opportunity to visit the University of Oxford, Northeastern University, NUS, and IIAI, among other institutions.
His research interests include computer vision, generative AI, spatial intelligence, world model, and embodied AI.
\end{IEEEbiography}

\vfill

\clearpage
\appendices

\section{Additional Ablation Study}
\label{sec:additional_ablation}

\noindent\textbf{Effect of multimodal perception.}
Tab.~\ref{tab:modal_increment} evaluates the contribution of geometric modalities through an incremental ablation.
Adding depth to text and RGB substantially improves both SR and SPL on R2R-CE and RxR-CE, while further incorporating point-cloud representations provides additional gains on both benchmarks.
Specifically, the multimodal configuration with depth and point clouds reaches 67.2 SR and 63.5 SPL on R2R-CE and 70.2 SR and 65.2 SPL on RxR-CE.
These results indicate that explicit geometric representations complement RGB appearance for spatially grounded navigation.
For reference, the complete MobileVLA-R1 configuration further reaches 68.3 SR and 65.2 SPL on R2R-CE and 71.5 SR and 66.8 SPL on RxR-CE.
\begin{table}[bhtp]
\centering
\small
\caption{
\textbf{Incremental ablation of multimodal perception on VLN-CE.}
The first three rows progressively add geometric modalities under the same training configuration.
The final row reports the complete MobileVLA-R1 configuration for reference.
}
\label{tab:modal_increment}
\resizebox{0.85\linewidth}{!}{
\begin{tabular}{lcccc}
\toprule
\multirow{2}{*}{Setting} & \multicolumn{2}{c}{R2R-CE} & \multicolumn{2}{c}{RxR-CE} \\
\cmidrule(lr){2-3} \cmidrule(lr){4-5}
& SR$\uparrow$ & SPL$\uparrow$ & SR$\uparrow$ & SPL$\uparrow$ \\
\midrule
Text + RGB & 62.5 & 59.0 & 66.0 & 61.5 \\
+ Depth    & 66.0 & 62.0 & 69.0 & 64.0 \\
+ Point Cloud & 67.2 & 63.5 & 70.2 & 65.2 \\
\midrule
MobileVLA-R1 (Full) & \textbf{68.3} & \textbf{65.2} & \textbf{71.5} & \textbf{66.8} \\
\bottomrule
\end{tabular}
}
\vspace{-0.4cm}
\end{table}

\noindent\textbf{Sensitivity to the movement reward weight.}
Tab.~\ref{tab:wm_sweep} analyzes the sensitivity to the movement reward weight $\lambda_{\rm mov}$ while keeping the remaining reward weights fixed.
Within the tested range, increasing $\lambda_{\rm mov}$ consistently improves navigation performance, with SR increasing from 65.2 to 68.3 and SPL from 61.0 to 65.2.
Meanwhile, the normalized contribution of $R_{\rm mov}$ gradually increases from 0\% to 32\%, while the behavior and format rewards together remain the dominant source of the reward-derived advantage.
This trend indicates that movement alignment provides a complementary optimization signal rather than replacing behavior correctness and structured-output validity.
Among the evaluated settings, $\lambda_{\rm mov}=1.0$ achieves the best overall performance.

\begin{table}[bhtp]
\centering
\small
\caption{
\textbf{Sensitivity to the movement reward weight on R2R-CE Val-Unseen.}
Results are obtained with deterministic action parsing.
We vary $\lambda_{\rm mov}$ while keeping $\lambda_{\rm beh}$ and $\lambda_{\rm fmt}$ fixed.
``Adv. Contrib.'' denotes the normalized contribution of each reward term to the reward-derived advantage, with all contributions summing to 100\%.
}
\label{tab:wm_sweep}
\resizebox{0.8\linewidth}{!}{
\begin{tabular}{cccccc}
\toprule
\multirow{2}{*}{$\lambda_{\rm mov}$} & \multirow{2}{*}{SR$\uparrow$} & \multirow{2}{*}{SPL$\uparrow$} & \multicolumn{3}{c}{Adv. Contrib. (\%)} \\
\cmidrule(lr){4-6}
& & & $R_{\rm mov}$ & $R_{\rm beh}$ & $R_{\rm fmt}$ \\
\midrule
0.00 & 65.2 & 61.0 &  0 & 62 & 38 \\
0.10 & 66.4 & 62.4 &  8 & 58 & 34 \\
0.25 & 67.1 & 63.5 & 16 & 54 & 30 \\
0.50 & 67.8 & 64.4 & 24 & 50 & 26 \\ \midrule
1.00 & \textbf{68.3} & \textbf{65.2} & 32 & 46 & 22 \\
\bottomrule
\end{tabular}
}
\vspace{-0.3cm}
\end{table}

\noindent\textbf{Effect of rationale source.}
Tab.~\ref{tab:teacher_ablation} evaluates different sources of reasoning supervision under the same training configuration.
Template-CoT consistently improves over No-CoT, while model-generated rationales provide further gains.
Among the evaluated sources, Gemini-CoT performs best, achieving an SR of 68.3 and an SPL of 65.2.
Because all variants use identical action targets, the performance differences primarily reflect the effect of rationale supervision rather than additional action annotations.
Overall, these results demonstrate that the source of supervised rationales has a measurable impact on navigation performance.

\begin{table}[bhtp]
\centering
\small
\caption{
\textbf{Ablation of rationale sources on R2R-CE Val-Unseen.}
All variants use identical action targets, backbone, structured output format, and training budget; only the source of supervised rationales is varied.
}
\label{tab:teacher_ablation}
\resizebox{0.6\linewidth}{!}{
\begin{tabular}{lcc}
\toprule
Rationale Source & SR$\uparrow$ & SPL$\uparrow$ \\
\midrule
No-CoT        & 64.0 & 59.6 \\
Template-CoT  & 65.1 & 60.8 \\
LLaMA-3-8B CoT& 66.2 & 62.0 \\
Gemini-CoT    & \textbf{68.3} & \textbf{65.2} \\
\bottomrule
\end{tabular}
}
\vspace{-0.3cm}
\end{table}

\noindent\textbf{Effect of policy optimization strategy.}
Tab.~\ref{tab:grpo_vs_ppo} compares PPO and GRPO under matched training settings.
GRPO improves SR from 64.1 to 68.3 and SPL from 60.4 to 65.2 on R2R-CE Val-Unseen, yielding gains of 4.2 and 4.8 points, respectively.
With reward definitions, regularization, output schema, and training budget controlled, the results indicate that group-relative optimization is more effective for the structured reasoning-to-control objective considered here.
\begin{table}[bhtp]
\centering
\small
\caption{
\textbf{Comparison of PPO and GRPO on R2R-CE Val-Unseen.}
Both methods use the same reward definitions, structured output schema, KL regularization, and training budget.
}
\label{tab:grpo_vs_ppo}
\resizebox{0.65\linewidth}{!}{
\begin{tabular}{lcc}
\toprule
Optimization Method & SR$\uparrow$ & SPL$\uparrow$ \\
\midrule
PPO  & 64.1 & 60.4 \\
GRPO & \textbf{68.3} & \textbf{65.2} \\
\bottomrule
\end{tabular}
}
\vspace{-0.4cm}
\end{table}

\begin{table}[bhtp]
\centering
\small
\caption{
\textbf{Comparison of optimization objectives on R2R-CE Val-Unseen.}
All variants use the same training data and action targets.
Reward-based variants use identical reward definitions and deterministic action parsing.
}
\label{tab:optimization_objective}
\resizebox{\linewidth}{!}{
\begin{tabular}{lccccc}
\toprule
Method & Sampling & Reward & Optimization Objective & SR$\uparrow$ & SPL$\uparrow$ \\
\midrule
SFT & \xmark & \xmark & Token-level CE & 58.0 & 53.2 \\
Reward-SFT & \xmark & \cmark & Reward-weighted CE & 62.4 & 58.3 \\
PPO & \cmark & \cmark & Policy Gradient & 64.1 & 60.4 \\
GRPO & \cmark & \cmark & Group-relative Objective & \textbf{68.3} & \textbf{65.2} \\
\bottomrule
\end{tabular}
}
\vspace{-0.3cm}
\end{table}

\noindent\textbf{Effect of optimization objective.}
Tab.~\ref{tab:optimization_objective} compares supervised and policy-based optimization objectives under matched training settings.
Reward-SFT consistently improves over standard SFT, while policy-based optimization provides further gains.
Among the evaluated objectives, GRPO performs best, achieving an SR of 68.3 and an SPL of 65.2, outperforming PPO by 4.2 and 4.8 points, respectively.
These results show that sequence-level reward optimization provides additional benefits beyond token-level reward-weighted supervision, and support the use of group-relative optimization for structured reasoning-to-control learning.

\section{CoT Data Generation and Quality Control}
\label{sec:supp_data_quality}

\noindent\textbf{Data split integrity.}
CoT annotations are generated exclusively from the official training splits of R2R, RxR, and QUARD.
No synthetic rationales or action annotations are generated for validation or test episodes.
All benchmark results are evaluated on the original validation splits using human-provided instructions and the corresponding official evaluation protocols.
During annotation generation, Gemini-2.5-Flash receives the task instruction, available observations, and state-action history, but is not provided with evaluation targets from any validation or test trajectory.
This separation prevents synthetic annotation generation from introducing evaluation-set supervision.

\begin{table*}[t]
\centering
\small
\caption{\textbf{Filtering statistics of \datasetname{}.}
``Ret.'' denotes the fraction of samples retained from the preceding stage.}
\label{tab:cot_filtering}
\resizebox{0.9\linewidth}{!}{
\begin{tabular}{lcccl}
\toprule
Filtering Stage & Input & Retained & Ret. & Primary Removal Reason \\
\midrule
Raw generation & 168K & 168K & 100.0\% & -- \\
Format validation & 168K & 158K & 94.0\% & malformed tags / missing fields \\
Action validation & 158K & 146K & 92.4\% & invalid commands / out-of-range values \\
Safety and relevance filtering & 146K & 139K & 95.2\% & unsafe or instruction-irrelevant outputs \\
Manual verification & 139K & 134K & 96.4\% & hallucination / action mismatch / visual inconsistency \\
\midrule
Final & 134K & 134K & 100.0\% & 18K episode / 78K step / 38K navigation \\
\bottomrule
\end{tabular}}
\vspace{-0.6cm}
\end{table*}
\noindent\textbf{Automatic filtering.}
The raw teacher generations are first processed using a sequence of automatic validity checks.
We enforce the structured \texttt{<think>...</think><answer>...</answer>} format and remove outputs with malformed tags, missing fields, or empty reasoning and answer segments.
For samples containing task-level actions, we additionally verify that continuous locomotion values are numerically valid and within the predefined ranges, and that discrete behavior labels belong to the valid behavior vocabulary.
Instruction-irrelevant and unsafe generations are subsequently removed.
These schema and action checks are used only for annotation quality control; physical execution of \modelname{} uses the learned
reasoning-conditioned action decoder rather than deterministic textual action parsing.

\noindent\textbf{Manual verification.}
After automatic filtering, the remaining annotations are manually examined for semantic consistency.
The verification focuses on three criteria:
(i) whether the rationale is consistent with the instruction and available observations;
(ii) whether the associated task-level action is compatible with the reasoning context and state-action history; and
(iii) whether the annotation contains hallucinated, visually inconsistent, or clearly erroneous content.
Samples violating any of these criteria are removed.
Deterministic annotation errors, such as formatting or unit normalization issues, are corrected when their intended values can be recovered unambiguously.
This stage removes approximately 5K additional samples, resulting in 134K annotations in the final dataset.

\noindent\textbf{Dataset composition.}
As summarized in Tab.~\ref{tab:cot_filtering}, the complete filtering pipeline reduces 168K raw teacher generations to 134K retained annotations.
The final dataset contains 18K episode-level, 78K step-level, and 38K navigation-level reasoning samples.
The different granularities provide complementary supervision ranging from long-horizon task interpretation to local decision making and executable task-level actions.

\noindent\textbf{Common annotation errors.}
The most frequent failures identified during automatic filtering include malformed structured outputs, missing action fields, and invalid continuous action values.
Manual verification further identifies hallucinated objects, inconsistencies between rationales and associated actions, and reasoning that is unsupported by the visual observations.
Removing these cases before training reduces noise in the reasoning supervision and ensures that retained annotations contain valid structured reasoning paired with well-defined action targets.

\end{document}